%% file: main.tex
\documentclass[runningheads]{llncs}

\usepackage{accv}

\usepackage{accvabbrv}
\usepackage{algorithm}
\usepackage{algpseudocode}
\usepackage{alltt,xcolor}
\usepackage[normalem]{ulem}
\usepackage{graphicx}
\usepackage{booktabs}
\usepackage{comment}
\usepackage{pifont}
\usepackage{multirow}
\usepackage{amsmath}
\newcommand{\xmark}{\ding{55}}%
\newcommand{\flowsam}{{\tt FlowI-SAM}}%
\newcommand{\motionsam}{{\tt FlowP-SAM}}%
\usepackage{mathrsfs}

\usepackage{graphicx}
\usepackage{booktabs}

\usepackage[accsupp]{axessibility}  % Improves PDF readability for those with disabilities.

\usepackage{hyperref}

\usepackage{orcidlink}

\begin{document}

\title{Moving Object Segmentation: \\ All You Need Is SAM (and Flow)}

\titlerunning{Moving Object Segmentation: All You Need Is SAM (and Flow)}

\author{Junyu Xie\inst{1}\orcidlink{0009-0002-1123-493X} \and
Charig Yang\inst{1}\orcidlink{0009-0003-7044-1901} \and
Weidi Xie\inst{1,2}\orcidlink{0009-0002-8609-6826} \and Andrew Zisserman\inst{1}\orcidlink{0000-0002-8945-8573}}

\authorrunning{J.~Xie et al.}

\institute{Visual Geometry Group, University of Oxford\and
School of Artificial Intelligence, 
Shanghai Jiao Tong University \\
\email{\{jyx,charig,weidi,az\}@robots.ox.ac.uk}
\href{https://www.robots.ox.ac.uk/~vgg/research/flowsam/}{\texttt{https://www.robots.ox.ac.uk/\textasciitilde vgg/research/flowsam/}}}

\maketitle
\input{sec/0-abstract}
\input{sec/1-introduction}

\input{sec/2-related_work}
\input{sec/3-method}

\input{sec/4-experiments}

\input{sec/5a-results}

\input{sec/5b-results}
\input{sec/6-conclusion}

\input{sec/Acknowledgement}

\bibliographystyle{splncs04}
\bibliography{egbib,vgg_local}

\clearpage
\input{supp_insert}

\end{document}

%% file: sec/0-abstract.tex
\begin{abstract} 
The objective of this paper is motion segmentation -- discovering and segmenting the moving objects in a video.
This is a much studied area with numerous careful, 
and sometimes complex, approaches and training schemes including: 
self-supervised learning, learning from synthetic datasets, 
object-centric representations, amodal representations,
and many more. Our interest in this paper is to determine if the Segment Anything model (SAM) can contribute to this task.

We investigate two models for combining SAM with optical flow that harness the segmentation power of SAM with the ability of flow to discover and group moving objects. 
In the first model, we adapt SAM to take optical flow, rather than RGB, as an input. In the second, SAM takes RGB as an input, and flow is used as a segmentation prompt. These surprisingly simple
methods, without any further modifications, outperform all previous
approaches by a considerable margin in both single and multi-object
benchmarks.  We also extend these frame-level segmentations to sequence-level segmentations that maintain object identity. 
Again, this simple model achieves outstanding performance across multiple moving object segmentation benchmarks.
\keywords{Moving Object Discovery \and Video Object Segmentation}
\end{abstract}

%% file: sec/1-introduction.tex
\section{Introduction}
\label{sec:intro}

\begin{figure}[!thbp]
    \centering
    \includegraphics[width=0.98\linewidth]{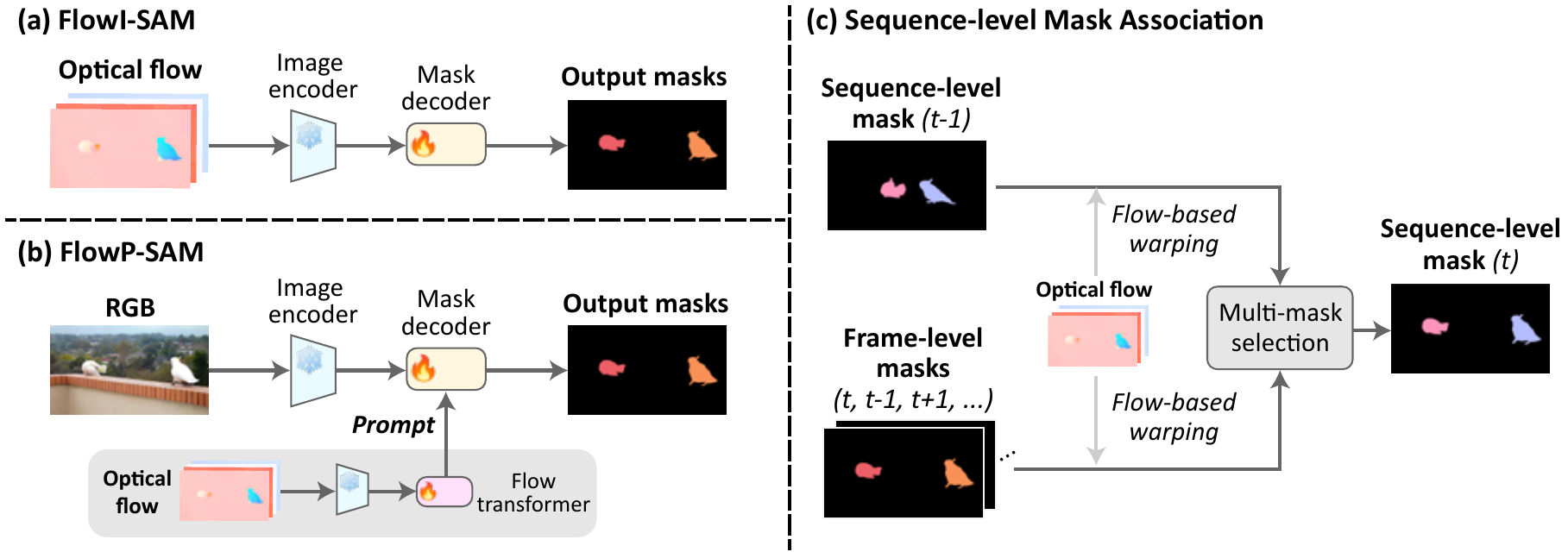}
    \vspace{-0.15cm}
    \caption{
    \textbf{Adapting SAM for Video Object Segmentation by incorporating flow.} \textbf{(a) Flow-as-Input}:~\flowsam~takes in optical flow \textit{only} and predicts frame-level segmentation masks. \textbf{(b) Flow-as-Prompt}:~\motionsam~takes in RGB and applies flow information as a prompt for frame-level segmentation. \textbf{(c) Sequence-level mask association}: as a post-processing step, the multi-mask selection module autoregressively transforms \textit{frame-level} mask outputs from \flowsam~and/or \motionsam~and produces \textit{sequence-level} masks in which object identities are consistent throughout the sequence.} 
    \label{fig:teaser}
\end{figure}

Recent research in image segmentation has been transformative, with the Segment Anything Model (SAM)~\cite{kirillov2023segment} emerging as a significant breakthrough. 
Leveraging large-scale datasets and scalable self-labelling, 
SAM enables flexible image-level segmentation across many scenarios~\cite{ma2024segment,wu2023medical,ren2024segment,tang2023sam,zhang2023samhelpsshadowwhen,chen2023sam}, facilitated by user prompts such as boxes, texts, and points. 
In videos, optical flow has played an important and successful role for {\em moving object segmentation} -- in that it can (i) discover moving objects, (ii) provide crisp boundaries for segmentation, and (iii) group parts of objects together if they move together. It has formed the basis for numerous methods of moving object discovery by self-supervised learning~\cite{Yang21a,meunier2023unsupervised,Xie22,Lamdouar21,meunier2022driven,Choudhury22,Safadoust23}. However, it faces segmentation challenges if objects are momentarily motionless, and in distinguishing foreground objects from background `noise'. This naturally raises the question: ``How can we leverage the power of SAM with flow for moving object segmentation in videos?''.

To this end, we explore two simple variants to effectively tailor SAM for motion segmentation. {\em First}, we introduce~\textit{\flowsam} (\cref{fig:teaser}a), an adaption of the original SAM that directly processes optical flow as a three-channel {\em input} image for segmentation, where points on a uniform grid are used as prompts. This approach leverages the ability of SAM to accurately segment moving objects against the static background, by exploiting the distinct textures and clear boundaries present in optical flow fields. However, it has less success in scenes where the optical flow arises from multiple interacting objects as the flow only contains limited information for separating them. 
{\em Second}, building on the strong ability of SAM on RGB image segmentation, we propose~\textit{\motionsam}~(\cref{fig:teaser}b) where the input is an RGB frame,
and flow guides SAM for moving object segmentation as {\em prompts}, produced by a trainable prompt generator. This method effectively leverages the ability of SAM on RGB image segmentation, with flow information acting as a selector of moving objects/regions within a frame. 
Additionally, we extend these methods from frame-level to {\em sequence-level} video segmentation~(\cref{fig:teaser}c) so that object identities are consistent throughout the sequence. We do this by introducing a matching module that auto-regressively chooses whether to select a new object or propagate the old one based on temporal consistency. 

In summary, this paper introduces and explores two models to leverage SAM for moving object segmentation in videos, enabling the principal moving objects to be discriminated from background motions. Our contributions are threefold:
\begin{itemize}
\setlength\itemsep{2pt}
    \item The~\flowsam~model,  which utilises optical flow as a three-channel input image for precise \textit{per-frame} segmentation and moving object identification.   
    \item The~\motionsam~model, a novel combination of dual-stream (RGB and flow) data, that employs optical flow to generate prompts, guiding SAM to identify and localise the moving objects in RGB images.
    \item New state-of-the-art unsupervised video object segmentation performance by a large margin on moving object segmentation benchmarks, including DAVIS16, DAVIS17-m, YTVOS18-m, and MoCA.
\end{itemize}

%% file: sec/2-related_work.tex
\section{Related Work}
\label{sec:related}

\par\noindent\textbf{Video Object Segmentation (VOS)} is an extensively studied task in computer vision. The objective is to segment the primary object(s) in a video sequence. Numerous benchmarks are developed for evaluating VOS performance, catering to both single-object \cite{Perazzi16, FliICCV2013, Ochs11, Lamdouar20} and multi-object \cite{Ponttuset17, Xu18} scenarios.
Two major protocols are widely explored in VOS research, namely unsupervised~\cite{Yang19,Yang21a,Lu_2019_CVPR,Ventura_2019_CVPR,cho2019key,Li_2018_CVPR,luiten2020unovost,lin2021video} and semi-supervised VOS~\cite{Vondrick18,Lai19,Lai20,Miao2022mamp,CVPR2019_CycleTime,jabri2020walk,caron2021emerging,iccv19_stm,yang2022deaot,cheng2022xmem, in-n-out}. 
Notably, the term ``unsupervised'' exclusively indicates that no groundtruth annotation is used \textit{during inference time} ({\em i.e.,}~no inference-time supervision). In contrast, the semi-supervised VOS employs first-frame groundtruth annotations to initiate the object tracking and mask propagation in subsequent frames. This paper focuses on unsupervised VOS and utilises motion as a crucial cue for object discovery. 

\vspace{3pt}
\par\noindent\textbf{Motion Segmentation} focuses on discovering objects through their movement and generating corresponding segmentation masks. Existing benchmarks for motion segmentation largely overlap with those used for VOS evaluation, especially in the single-object case. 
For multi-object motion segmentation, datasets~\cite{Xie22, xie2023appearancebased} have been specifically curated from VOS benchmarks to exclusively focus on sequences with dominant locomotion. 
There are two major setups in the motion segmentation literature: one that relies on motion information \textit{only} to distinguish moving elements from the background through spatial clustering~\cite{Yang21a, meunier2022driven, meunier2023unsupervised} or explicit supervision~\cite{Lamdouar21, Xie22}; the other~\cite{Bideau16,Mahendran_2018_ECCV_Workshops, Yang_2021_CVPR,Choudhury22,xie2023appearancebased, 10030403} that enhances motion-based object discovery by incorporating appearance information. We term these two approaches ``flow-only'' and ``RGB-based'' segmentation, respectively, and explore both setups in this work.

\vspace{3pt}
\par\noindent\textbf{Segment Anything Model (SAM)}~\cite{kirillov2023segment} has demonstrated impressive ability on image segmentation across various scenarios. It was trained on the SA-1B datasets with over one billion self-labelled masks and $11$ million images. Such large-scale training renders it a strong zero-shot generalisability to unseen domains. Many works adapt the SAM model to perform different tasks, such as tracking \cite{cheng2023segment}, change detection \cite{zheng2024segment}, and 3D segmentation~\cite{cen2023saga}. 
Some other works extend SAM towards more efficient models \cite{mobile_sam,zhao2023fast,xiong2023efficientsam}, and more domains \cite{ma2024segment,wu2023medical,ren2024segment,chen2023sam}.
However, most studies follow the default prompt options in SAM (\emph{i.e.,} points, boxes, and masks). Recent works~\cite{zou2023segment,Sun_2024_CVPR} have shown that more versatile prompts, including scribbles and visual references, can lead to improvements. In this paper, we explore a novel route that prompts SAM with optical flow and demonstrate its effectiveness for moving object segmentation.

%% file: sec/3-method.tex
\section{SAM Preliminaries}
\label{sec:sampre}

The Segment Anything Model (SAM) is engineered for high-precision image segmentation, accommodating both user-specified prompts and a fully autonomous operation mode. 
When guided by user input, SAM accepts various forms of prompts including points, boxes, masks, or textual descriptions to accurately delineate the segmentation targets. Alternatively, in its automatic mode, SAM uses points on a uniform grid as prompts, to propose all plausible segmentation masks that capture objects and their hierarchical subdivisions—objects, parts, and subparts.
In this case, the inference is repeated for each prompt of the grid, generating masks for each prompt in turn, and the final mask selection is guided by the predicted mask IoU scores.

Architecturally, SAM is underpinned by three foundational components:
(i) Image encoder extracts strong image features via a heavy Vision Transformer (ViT) backbone, which is pre-trained by the Masked Auto-Encoder (MAE) approach; (ii) The prompt encoder converts the input prompts into positional information which helps with locating the segmentation target; (iii) Mask decoder features a light-weight two-way transformer that takes in a combination of encoded prompt tokens, learnable mask tokens, and an IoU prediction token as input queries. These queries iteratively interact with the dense spatial features from image encoder, leading to the final mask predictions and IoU estimations.
In the next sections, we describe two distinct, yet simple variants to effectively tailor SAM for motion segmentation.

\section{Frame-Level Segmentation I: Flow as Input} 
\label{sec:frame1}

\begin{figure}[t!]
    \centering
    \includegraphics[width=0.98\linewidth]{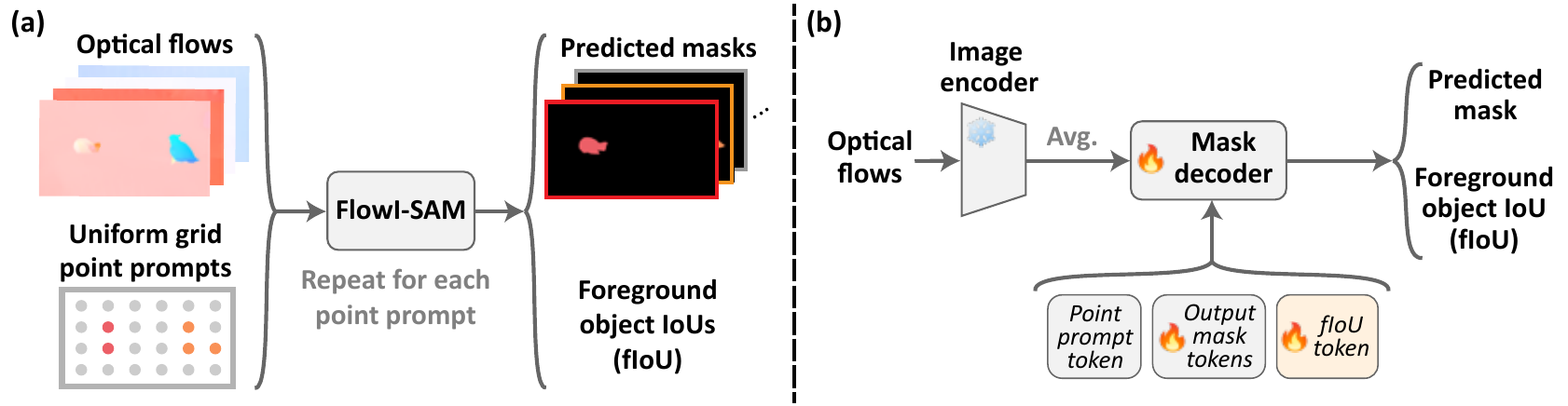}
    \vspace{-0.15cm}
    \caption{\textbf{Overview of~\flowsam.} \textbf{(a)} Inference pipeline of~\flowsam. \textbf{(b)} Architecture of~\flowsam~with trainable parameters labelled. The point prompt token is generated by a frozen prompt encoder.}
    \label{fig:motion_seg}
\end{figure}

\noindent In this section, we focus on discovering moving objects from individual frames by exploiting motion information \textit{only}, to yield corresponding segmentation masks.
Formally, given the optical flow input $F_t \in \mathbb{R}^{H \times W \times 3}$  at frame $t$, 
we aim to predict a segmentation mask ${M^i_t} \in \mathbb{R}^{H \times W}$ together with a foreground object IoU (fIoU) score $s_{\text{fIoU},t}^i \in \mathbb{R}^{1}$ for each object $i$,
\begin{equation}
    \{M^i_t, s_{\text{fIoU},t}^i\}_{i=0}^{N} = \Phi_{\text{\flowsam}}(F_t)
\end{equation}
To adapt SAM for this new task, we formulate \flowsam~($\Phi_{\text{\flowsam}}$) by finetuning it on optical \textbf{Flow} \textbf{I}nputs, and {re-purpose} the original IoU prediction head to instead predict the fIoU, as illustrated in~\cref{fig:motion_seg}b. By definition, fIoU is a scalar that measures the ``objectness'': fIoU is $0$ if the mask belongs to the background, 
and equal to the IoU between predicted and GT object masks for foreground moving object masks. A high fIoU indicates the predicted mask corresponds to the entire object, while a low fIoU might suggest the mask is erroneous or only captures a small part of the object.  

\vspace{3pt}
\par{\noindent \textbf{Flow Inputs with Multiple Frame Gaps.}} 
To mitigate the effect of noisy optical flow, 
{\em i.e.}~complicated flow fields due to stationary parts, articulated motion, and object interactions, 
{\em etc.}, we consider multiple flow inputs $\{F_{t,g}\}$ with different frame gaps ({\em e.g.,}~$g \in \{(1,$-$1),(2,$-$2) \}$) for both training and evaluation stages. These multi-gap flow inputs are \textbf{independently} processed by the image encoder to obtain dense spatial features $\{d_{t,g}\}$ at a lower resolution $h \times w$, which are then combined by averaging the spatial feature maps across different flow gaps, {\em i.e.,}~$d_t = \text{Average}_g(\{d_{t,g}\}) \in \mathbb{R}^{h \times w \times d}$. These averaged spatial features are then treated as keys and values in the mask decoder.

\vspace{3pt}
\par{\noindent \textbf{\flowsam~Inference.}}  
To discover all moving objects from flow input, 
the \flowsam~model is prompted by points on a uniform grid. Each point prompt outputs a pair of mask and objectness score predictions. This mechanism is the same as in the original SAM formulation, and is illustrated in~\cref{fig:motion_seg}a. The final segmentation is selected using Non-Maximum Suppression (NMS) based on the predicted fIoU and overlap ratio.

\vspace{3pt}
\par{\noindent \textbf{\flowsam~Training.}} 
To adapt the pre-trained SAM model for optical flow inputs, 
we finetune the lightweight mask decoder, 
while the image encoder and the prompt encoder remain frozen. 
The overall loss is formulated as:
\begin{equation}
    \mathcal{L}_{\text{\flowsam}} = \frac{1}{NT}\sum_{i,t}^{N,T}\left(\mathcal{L}_{\text{BCE}}(M^i_t, \hat{M}^i_t) +  \lambda_f \lVert
         s_{\text{fIoU},t}^i -  \hat{s}_{\text{fIoU},t}^i \lVert^2 \right)
\end{equation}
where $\hat{M}^i_t$ and $\hat{s}_{\text{fIoU},t}^i$ denote the groundtruth segmentation masks and fIoU, and $\lambda_f$ is a scale factor.

\section{Frame-Level Segmentation II: Flow as Prompt} 
\label{sec:frame2}
\begin{figure}[t!]
    \centering
    \includegraphics[width=0.98\linewidth]{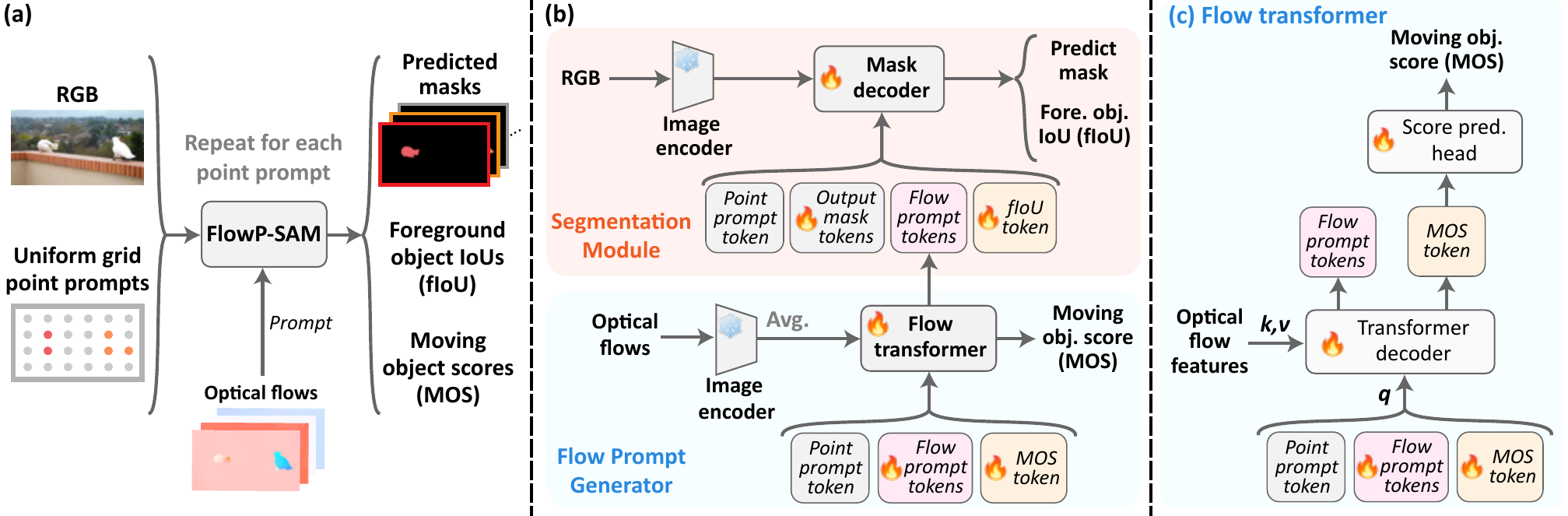}
    \vspace{-0.15cm}
    \caption{\textbf{Overview of~\motionsam.} \textbf{(a)} Inference pipeline of~\motionsam. \textbf{(b)} Architecture of~\motionsam. The flow prompt generator produces flow prompts to be injected into a SAM-like RGB-based segmentation module. Both modules take in the same point prompt token, which is obtained from a frozen prompt encoder. \textbf{(c)} Detailed architecture of the flow transformer. The input tokens function as queries within a lightweight transformer decoder, iteratively attending to dense flow features. 
    The output moving object score (MOS) token is then processed by an MLP-based head to predict a score indicating whether the input point prompt corresponds to a moving object.}
    \label{fig:motionrgb_seg}
\end{figure}

In this section, we adapt SAM for video object segmentation by processing RGB frames, with optical flow as a prompt. 
We term this frame-level segmentation architecture~\motionsam~for \textbf{Flow} as \textbf{P}rompt SAM. 
As shown in~\cref{fig:motionrgb_seg}b, \motionsam~encompasses two major modules, 
namely the flow prompt generator and the segmentation module.
The flow prompt generator takes optical flow as inputs, and produces flow prompts that can be used as supplemental queries to infer frame-level segmentation masks ${M^i_t}$ from RGB inputs $I_t$. Formally,
\begin{equation}
    \{M^i_t, \, s_{\text{fIoU},t}^i, \, s_{\text{MOS},t}^i\}_{i=0}^{N} = \Phi_{\text{\motionsam}}(F_t, I_t)
\end{equation}
where $s_{\text{MOS},t}^i$ indicates the moving object score (MOS) predicted by the flow prompt generator, while $s_{\text{fIoU},t}^i$ denotes the foreground object IoU (fIoU) estimated by the segmentation module.
Specifically, MOS measures whether each input point prompt (therefore the resultant mask) is within a moving object region based on observing flow fields. Groundtruth MOS scores are binary (\emph{i.e.,} $\hat{s}_{\text{MOS},t}^i = 1$ if the point prompt is within GT annotation, and $\hat{s}_{\text{MOS},t}^i = 0$ otherwise).
On the other hand, fIoU follows the same formulation as in~\flowsam, {\em i.e.,}~predicting IoUs for foreground objects and yielding $0$ for background regions.

\vspace{3pt}
\par{\noindent \textbf{Flow Prompt Generator}} consists of (i) a frozen SAM image encoder, where the dense spatial features are extracted from optical flow inputs at different frame gaps, followed by an averaging across frame gaps; 
(ii) a flow transformer, with the detailed architecture depicted in~\cref{fig:motionrgb_seg}c, where we first stack the input point prompt ({\em i.e.,} a positional embedding)
with learnable flow prompts and moving object score (MOS) tokens to form queries. 
These queries then iteratively attend to dense flow features in a lightweight transformer decoder. 
There are two outputs from the flow prompt generator, namely, the refined flow prompts, and an MOS token, which is subsequently processed by an MLP-based head to yield a final moving object score.

\vspace{3pt}
\par{\noindent \textbf{Segmentation Module}} has a structure that resembles the original SAM, except for two adaptions: (i) The IoU-prediction head is re-purposed to predict foreground object scores (fIoU)~(same as the~\flowsam); (ii) The outputs tokens from flow prompt generator are injected as additional query inputs. 

\vspace{3pt}
\par{\noindent \textbf{\motionsam~Inference.}} 
Similar to~\flowsam, we prompt ~\motionsam~ with single point prompts from a uniform grid to iteratively predict possible segmentation masks, together with {MOS and fIoU} estimations. These predicted scores are averaged, {\em i.e.,}~$(s_{\text{MOS},t}^i + s_{\text{fIoU},t}^i)/2$, and then utilised to guide post-processing which involves the NMS and overlay of selected masks.

\vspace{3pt}
\par{\noindent \textbf{\motionsam~Training.}} We train~\motionsam~in an end-to-end fashion while keeping the SAM pre-trained prompt encoder and image encoders frozen. 
The {flow transformer} is trained from scratch, 

\begin{align}
    \mathcal{L}_{\text{\motionsam}} = &\frac{1}{NT}\sum_{i,t}^{N,T}\Big(\Big.\mathcal{L}_{\text{BCE}}(M^i_t, \hat{M}^i_t)   \nonumber \\
    + &  \lambda_m 
    \mathcal{L}_{\text{BCE}}(s_{\text{MOS},t}^i, \hat{s}_{\text{MOS},t}^i) +\lambda_f \lVert
         s_{\text{fIoU},t}^i - \hat{s}_{\text{fIoU},t}^i \lVert^2 \Big)\Big.
\end{align}
where $\hat{M}^i_t$ corresponds to the groundtruth mask. $\hat{s}_{\text{MOS},t}^i$ and $\hat{s}_{\text{fIoU},t}^i$ indicate the groundtruth of two predicted scores, with $\lambda_m$ and $\lambda_f$ being the scale factors.

\section{Sequence-level Mask Association}
\label{sec:temp}

In this section, we outline our method for linking the frame-wise predictions for each moving object into a continuous track throughout the sequence. Specifically, we compute two types of masks: {\em frame-wise} masks $M$  at the current frame using the model (\flowsam~and/or \motionsam); and {\em sequence-level} masks $\mathscr{M}$, that are obtained by propagating the initial frame prediction with optical flow, we then update the mask of the current frame by making a comparison between the two.
The following section details our update mechanism.

\vspace{3pt}
\par \noindent \textbf{Update Mechanism.} 
This process aims to associate object masks across frames, as well as to determine whether the sequence-level results at a particular frame should be obtained directly from frame-level predictions at that frame or by propagating from previous frame results.

Specifically, given a sequence-level mask for object $i$ at frame $t-1$ ({\em i.e.,} $\mathscr{M}^i_{t-1}$), 
we first warp it to frame $t$ using optical flow, 
\begin{equation}
    \mathscr{M}^i_{t\leftarrow t-1} = \text{warp}(\mathscr{M}^i_{t-1}, F_{t-1})
\end{equation}

We then consider three sets of masks:
(i) the warped masks $\{\mathscr{M}^i_{t\leftarrow t-1}\}$; (ii) the frame-level predictions $\{{M}^i_{t}\}$ at frame $t$; and (iii) the frame-level predictions from neighboring frames (with $\Delta{t}$ gap) after aligning them to the current frame using optical flow ({\em i.e.,} $\{{M}^i_{t\leftarrow t+\Delta{t}}\}$). 
For each pair of mask sets, we perform a pairwise Hungarian matching based on the IoU score, resulting in three pairings in total.
The Hungarian matched pairs can then reflect the \textit{temporal consistency} across these predictions based on the \textit{transitivity} principle, \ie if object $i$ in (i) matches with object $j$ in (ii) and object $k$ in (iii), then the latter two objects must also match with each other. If such transitivity holds, we set the consistency score $c_i = 1$, and $c_i = 0$ otherwise.

This matching process is repeated for $\Delta{t} \in \{1,2,-1,-2\}$, resulting in an averaged consistency score $\bar{c}_i$, which guides the following mask update:
\begin{equation}
\mathscr{M}^i_{t}  = \begin{cases} 
    {M}^i_{t} & \text{if } \bar{c}_i \geq 0.5 \\
    \mathscr{M}^i_{t\leftarrow t-1}  & \text{if } \bar{c}_i < 0.5 
\end{cases}
\label{eq:warp}    
\end{equation}
where $\mathscr{M}^i_{t}$ denotes the sequence-level mask prediction for object $i$ at frame $t$.

The rationale behind this is that the two methods have their own strengths and drawbacks: propagation is safe in preserving the object identity, but the mask quality degrades over time, whereas updating ensures high-quality masks but comes with the risk of mis-associating object identities. Thus, if the current frame-wise mask is  temporally consistent, then we can be reasonably confident to update the mask, if not, then we choose the safer option and propagate the previous mask.

Note, we do this separately for each object $i \in N$, which gets updated or propagated independently. We lastly layer all objects back together (to its original order) and remove any overlaps to obtain the final sequence-level predictions.

%% file: sec/4-experiments.tex
\section{Experiments}
\label{sec:experiment}

\subsection{Datasets}
\vspace{3pt}
\par{\noindent \textbf{Single-Object Benchmarks.}}
For single-object motion segmentation, we adopt standard datasets, 
including DAVIS2016~\cite{Perazzi16}, SegTrackv2~\cite{FliICCV2013}, FBMS-59~\cite{OB14b}, and MoCA~\cite{Lamdouar20}. Although SegTrackv2 and FBMS-59 include a few multi-object sequences, 
following the common practice~\cite{Yang21a, Lamdouar21}, we treat them as single-object benchmarks by grouping all moving objects into a single foreground mask. MoCA stands for Moving Camouflaged Animals, designed as a camouflaged object detection benchmark. Following~\cite{Yang21a, Lamdouar21, Xie22}, we adopt a filtered MoCA dataset by excluding videos with predominantly no locomotion.

\vspace{3pt}
\par{\noindent \textbf{Multi-Object Benchmarks.}}
In terms of multi-object segmentation, we report the performance on DAVIS2017~\cite{Ponttuset17}, DAVIS2017-motion~\cite{Ponttuset17, Xie22}, and YouTube-VOS2018-motion~\cite{Xu18, xie2023appearancebased}, where DAVIS2017 is characterised by predominantly moving objects, {each annotated as distinct entities. For example, a man riding a horse would be separately labelled.} 
In contrast, DAVIS2017-motion re-annotates the objects based on their joint movements such that objects with shared motion are annotated as a single entity.
For example, a man riding a horse is annotated as a single entity due to their shared motion.

The YouTubeVOS2018-motion~\cite{xie2023appearancebased} dataset is a curated subset of the original YouTubeVOS2018~\cite{Xu18}. It specifically excludes video sequences involving common motion, severe partial motion, and stationary objects, making it ideally suited for motion segmentation evaluation. Whereas, the original dataset also annotates many stationary objects and only provides partial annotations for a subset of moving objects.

\vspace{3pt}
\par{\noindent \textbf{Summary of Evaluation Datasets.}} To investigate the role of motion in object discovery and segmentation, we adopt all aforementioned benchmarks, which consist of only predominantly moving object sequences. Notably, for the evaluation of~\flowsam, we exclude the class-labelled DAVIS17 dataset, as commonly moving objects cannot be separated solely based on motion cues.

\vspace{3pt}
\par{\noindent \textbf{Training Datasets.}} 
To adapt the RGB pre-trained SAM for moving object discovery and motion segmentation, we train both~\flowsam~and~\motionsam~first on the synthetic dataset introduced by~\cite{Xie22}, as described below, and then on real-world video datasets, including DAVIS16, DAVIS17, and DAVIS17-m.

\vspace{3pt}
\subsection{Evaluation Metrics}
\label{subsec:evalmetric}
To assess the accuracy of predicted masks, we report intersection-over-union ($\mathcal{J}$), except for MoCA where only the ground-truth bounding boxes are given, we instead follow the literature~\cite{Yang21a} and report the detection success rate (SR). Regarding multi-object benchmarks, we additionally report the contour accuracy ($\mathcal{F}$) in \Cref{supsec:quantitative}.

In this work, we differentiate between the frame-level and sequence-level methods, 
and adopt two distinct evaluation protocols: 
(i) Since frame-level methods generate the segmentation \textit{independently} for each frame, 
we apply per-frame Hungarian matching to match the object masks between predictions with the groundtruth, before the evaluation; (ii) Conversely, sequence-level methods employ an extra step to link object masks across frames. As a result, the Hungarian matching is conducted globally for each sequence, 
{\em i.e.,} the object IDs between predicted and groundtruth masks are matched once per sequence. 
Given the added complexity and potential errors during frame-wise object association, sequence-level predictions are often considered a greater challenge.

\vspace{3pt}
\subsection{Implementation Details}
In this section, we summarise the experimental setting in our frame-level segmentation models. For more information regarding detailed architectures, hyperparameter settings, and sequence-level mask associations, please refer to~\Cref{supsec:impl}.

\vspace{3pt}
\par{\noindent \textbf{Flow Computation.}} We adopt an off-the-shelf method (RAFT~\cite{Teed20}) to estimate optical flow with multiple frame gaps at $(1,$-$1)$ and $(2,$-$2)$, except for YTVOS18-m and FBMS-59, where higher frame gaps at $(3,$-$3)$ and $(6,$-$6)$ are used to compensate for slow motion. Following common practice~\cite{Yang21a, Xie22}, we convert optical flow into 3-channel RGB format using a standard color wheel~\cite{colorwheel}.

\vspace{3pt}
\par{\noindent \textbf{Model Settings.}} 
For both~\flowsam~and~\motionsam, we follow the default SAM setting and adopt the first output mask token (out of four) for mask predictions.  For~\flowsam, we deploy two versions of the pre-trained SAM image encoder, specifically ViT-B and ViT-H, to extract optical flow features. 

Regarding~\motionsam, for efficiency reasons, we utilise ViT-B to encode optical flows and employ ViT-H as the image encoder for RGB frames. We initialise the flow prompt generator with $4$ learnable flow prompt tokens, which are subsequently processed by a light-weight two-layer transformer decoder in the flow transformer module.

\vspace{3pt}
\par{\noindent \textbf{Evaluation Settings.}} 
At inference time, for~\flowsam~with flow-only inputs, we input independent point prompts over a $10 \times 10$ uniform grid, while for~\motionsam, to take account for more complicated RGB textures, we consider a large grid size of $20 \times 20$. 

\vspace{3pt}
\par{\noindent \textbf{Mask Selection over Multiple Point Prompts.}} During post-processing, we utilise the predicted scores (fIoU for~\flowsam, and an average of MOS and fIoU for~\motionsam) as guidance throughout the mask selection process: (i) Non-Maximum Suppression (NMS) is applied to filter out repeating masks and keep the ones with higher scores; (ii) The remaining masks are then ranked according to their scores and top-$n$ masks are retained ($n=5$ for~\flowsam, and $n=10$ for~\motionsam); (iii) These $n$ masks are overlaid by allocating masks with higher scores at the front. 

\vspace{3pt}
\par{\noindent \textbf{Training Settings.}} 
The training is performed in two stages, which involves synthetic pre-training on the dataset proposed by~\cite{Xie22}, followed by finetuning on the real DAVIS sequences, as detailed in~\Cref{supsec:impl_framelevel}.
YTVOS is not used for fine-tuning as there is only a low proportion of moving object sequences.
We train both models in an end-to-end manner using the Adam Optimiser at a learning rate of $3e^{-5}$. The training was conducted on a single NVIDIA A40 GPU, with each mode taking roughly $3$ days to reach full convergence.

\vspace{3pt}
\subsection{Ablation Study}
In this section, we present a series of ablation studies on key designs in the per-frame~\motionsam~model. We refer the reader to~\Cref{supsec:ablation} for a more detailed ablation analysis on the designs in~\flowsam~and~\motionsam~models, as well as in our sequence-level method.

\begin{table}[!t]
\centering
\setlength\tabcolsep{7pt}
\resizebox{\textwidth}{!}{
\begin{tabular}{cccccc}  
\toprule
\multirow{2}{*}{Stage} & Predicted scores & Flow & FT mask & DAVIS17 & DAVIS16 \\
& for post-processing & prompt & decoder & $\mathcal{J}$ $\uparrow$ & $\mathcal{J}$ $\uparrow$\\
\midrule
 & IoU & \xmark  & \xmark & $25.2$  & $30.3$  \\
\midrule
\multirow{2}{*}{+}~\multirow{2}{*}{\shortstack{Train flow prompt\\generator}}  & IoU & \textcolor{blue}{$\checkmark$}  & \xmark & $61.9$  & $80.6$  \\
% \cmidrule(r){2-6}
  & (\textcolor{blue}{MOS}+IoU)/2 & $\checkmark$  & \xmark & $63.7$  & $81.4$  \\
\midrule
\multirow{2}{*}{+}~\multirow{2}{*}{\shortstack{Finetune segment\\-ation module}}  & (MOS+IoU)/2 & $\checkmark$  & \textcolor{blue}{$\checkmark$} & $65.5$  & $81.5$  \\
 & (MOS+\textcolor{blue}{fIoU})/2 & $\checkmark$ & $\checkmark$ & $69.9$  & $86.1$  \\
\bottomrule
\end{tabular}}
\vspace{0.15cm}
\caption{\textbf{Ablation analysis of~\motionsam.} The study starts from the vanilla SAM checkpoint and progressively introduces new components (labelled in \textcolor{blue}{blue}). ``MOS'' is short for the moving object score, and ``fIoU'' indicates the foreground object IoU. The results are shown for frame-level predictions.}
\label{tab:abla_flowpsam}
\end{table}

\vspace{3pt}
\par{\noindent \textbf{Ablation Studies for~\motionsam.}} 
As illustrated in~\Cref{tab:abla_flowpsam}, we start from the vanilla SAM and progressively add our proposed components. Note that, we adopt the same inference pipeline ({\em i.e.,}~the same point prompts and post-processing steps) for all predictions shown. Since the {foreground object IoU (fIoU)} is not predicted by the vanilla SAM, we instead apply default IoU predictions to guide the mask selection.

We \textit{train the flow prompt generator} to simultaneously predict flow prompt tokens and moving object scores (MOS). 
% \weidi{this is flow prompt generator, right ?}
The injection of flow prompts into the standard RGB-based SAM architecture results in notable enhancements, verifying the value of motion information for accurately determining object positions and shapes. Additionally, employing {MOS} as additional post-processing guidance yields further improvements.

Upon \textit{finetuning the segmentation module}, we observe a slight enhancement in performance. Finally, substituting the default IoU predictions with {fIoU} scores achieves more precise mask selection, as evidenced by the improved results.

\vspace{3pt}
\par{\noindent \textbf{Discussion on the effectiveness of MOS and fIoU.}}
As outlined in Sect.~\ref{sec:sampre}, the original SAM framework over-segments images into objects, parts, and subparts, which cannot be further distinguished using the default SAM IoU estimations. 
To adapt this setup for \emph{object-only} discovery, we propose two new scores, \emph{i.e.,} MOS and fIoU, as new criteria to filter out non-object masks. These scores effectively assess the ``objectness'' of masks: MOS determines if the predicted masks represent moving objects, while fIoU evaluates whether the masks depict complete objects and are not background segments. The effectiveness of this adaptation is validated in Table~\ref{tab:abla_flowpsam}, where replacing IoU estimations with MOS + fIoU scores leads to noticeable performance boosts.

%% file: sec/5a-results.tex
\subsection{Quantitative Results}
\label{sec:results}

Given the distinct evaluation protocols outlined in~\cref{subsec:evalmetric}, 
we report our method separately, with a frame-level analysis for~\flowsam~(introduced in~\cref{sec:frame1}) and~\motionsam~(introduced in~\cref{sec:frame2}), followed by a sequence-level evaluation.

\begin{table}[!t]
\centering
\setlength\tabcolsep{3pt}
\resizebox{\textwidth}{!}{
\begin{tabular}{lccccccccc}  
\toprule
 &   & & \multicolumn{3}{c}{Multi-object benchmarks} & \multicolumn{4}{c}{Single-object benchmarks} \\ 
\cmidrule(lr){4-6}
\cmidrule(lr){7-10}
{} & 
&  & \multirow{2}{*}{\shortstack{YTVOS18-m \\$\mathcal{J}$ $\uparrow$}}  & \multirow{2}{*}{\shortstack{DAVIS17-m \\$\mathcal{J}$ $\uparrow$}}  & \multirow{2}{*}{\shortstack{DAVIS17 \\$\mathcal{J}$ $\uparrow$}}   & \multirow{2}{*}{\shortstack{DAVIS16 \\$\mathcal{J}$ $\uparrow$}} & \multirow{2}{*}{\shortstack{STv2 \\$\mathcal{J}$ $\uparrow$}}  & \multirow{2}{*}{\shortstack{FBMS\\$\mathcal{J}$ $\uparrow$}}  & \multirow{2}{*}{\shortstack{MoCA\\SR $\uparrow$}} \\ 

Model  & Flow & RGB  &   &  & &  & & \\
\midrule
\multicolumn{6}{l}{\textit{\textbf{Flow-only methods}}} \\
\midrule
COD~\cite{Lamdouar20} & $\checkmark$  & \xmark &  $-$  &  $-$  & $-$ &  $65.3$ & $-$ & $-$ & $0.236$ \\
$^{\dag}$MG~\cite{Yang21a} & $\checkmark$ & \xmark  & $37.0$  & $38.4$  &  $-$  & $68.3$ & $58.6$ & $53.1$ & $0.484$ \\
$^{\dag}$EM~\cite{meunier2022driven} & $\checkmark$ & \xmark &  $-$   & $-$  & $-$  &   $69.3$ &  $55.5$  &  $57.8$ & $-$ \\
\textbf{\flowsam~(ViT-B)} & $\checkmark$ & \xmark  & $56.7$ & $63.2$  &  $-$  &  $\mathbf{79.4}$ & $69.0$ & $72.9$ & $\mathbf{0.628}$  \\
\textbf{\flowsam~(ViT-H)} & $\checkmark$ & \xmark & $\mathbf{58.6}$ & $\mathbf{65.7}$   &  $-$  & $79.1$ & $\mathbf{70.1}$ & $\mathbf{75.1}$ &  $0.625$ \\
\midrule
\multicolumn{6}{l}{\textit{\textbf{RGB-based methods}}} \\
\midrule
$^{\dag}$VideoCutLER~\cite{wang2023videocutler} & \xmark & $\checkmark$ & $59.0$  & $57.4$   & $41.7$ & $-$ & $-$ & $-$ & $-$  \\
$^{\dag}$Safadoust et al.~\cite{Safadoust23} & \xmark & $\checkmark$  & $-$  & $59.3$   & $-$& $-$ & $-$ & $-$ & $-$  \\
MATNet~\cite{zhou20} & $\checkmark$ & $\checkmark$  & $-$ & $-$ & $56.7$  & $82.4$ & $50.4$ & $76.1$ & $0.544$ \\
DystaB~\cite{Yang_2021_CVPR} & \xmark & $\checkmark$  & $-$ & $-$ & $-$ &  $82.8$ &  ${74.2}$ & $75.8$  & $-$ \\
AMC-Net~\cite{amcnet} & $\checkmark$ & $\checkmark$  & $-$ & $-$ & $-$ &  $84.5$  &  $-$   &  $76.5$  & $-$ \\
TransportNet~\cite{transportnet} & $\checkmark$ & $\checkmark$  & $-$ & $-$ & $-$ & $84.5$ & $-$ & $78.7$ & $-$ \\
TMO~\cite{tmo} & $\checkmark$ & $\checkmark$  & $-$ & $-$ & $-$ & ${85.6}$  &  $-$  & ${79.9}$ & $-$ \\
\textbf{\motionsam} & $\checkmark$ & $\checkmark$   & ${76.9}$ & ${78.5}$ & ${69.9}$ &  ${86.1}$ & ${83.9}$ & ${87.9}$ & $\mathbf{0.645}$  \\
\textbf{\motionsam+\flowsam} & $\checkmark$ & $\checkmark$   &  $\mathbf{77.4}$ & $\mathbf{80.0}$ & $\mathbf{71.6}$  & $\mathbf{86.2}$  & $\mathbf{84.2}$ & $\mathbf{88.7}$   &  $\mathbf{0.645}$  \\
\bottomrule
\end{tabular}}
\vspace{0.15cm}
\caption{
\textbf{Frame-level comparison on video object segmentation benchmarks.} ``${\dag}$'' indicates models that are trained without human annotations. For the results in the last row, we combine frame-level predictions from~\motionsam~and~\flowsam~(ViT-H). 
}
\label{tab:vos_frame}
\end{table}

\vspace{3pt}
\par{\noindent \textbf{Frame-Level Performance.}} Table \ref{tab:vos_frame} distinguishes between flow-only and RGB-based methods, where the former adopts optical flow as the only input modality, and the latter takes in RGB frames with optional flow inputs. Note that, the performance for some recent self-supervised methods is also reported, owing to the lack of the supervised baselines.

For flow-only segmentation, our~\flowsam~(with both SAM image encoders) outperforms the previous methods by a large margin (>$10\%$). For RGB-based segmentation, our~\motionsam~also achieves state-of-the-art performance, particularly excelling at multi-object benchmarks. By combining these two frame-level predictions (\flowsam+\motionsam), 
we observe further performance boosts. This suggests the complementary roles of the flow and RGB modalities in frame-level segmentation, particularly when there are multiple moving objects involved.  
In particular, we show that using both models in tandem by layering \flowsam's predictions behind that of \motionsam~allows the model to fill in on missed predictions (such as motion blur, poor lighting, or small objects).

\begin{table}[!t]
\centering
\setlength\tabcolsep{3pt}
\resizebox{\textwidth}{!}{
\begin{tabular}{lccccccccc}  
\toprule
 &   & & \multicolumn{3}{c}{Multi-object benchmarks} & \multicolumn{4}{c}{Single-object benchmarks} \\ 
\cmidrule(lr){4-6}
\cmidrule(lr){7-10}
{} & 
&  & \multirow{2}{*}{\shortstack{YTVOS18-m \\$\mathcal{J}$ $\uparrow$}}  & \multirow{2}{*}{\shortstack{DAVIS17-m \\$\mathcal{J}$ $\uparrow$}} & \multirow{2}{*}{\shortstack{DAVIS17 \\$\mathcal{J}$ $\uparrow$}}    & \multirow{2}{*}{\shortstack{DAVIS16 \\$\mathcal{J}$ $\uparrow$}} & \multirow{2}{*}{\shortstack{STv2 \\$\mathcal{J}$ $\uparrow$}}  & \multirow{2}{*}{\shortstack{FBMS\\$\mathcal{J}$ $\uparrow$}}  & \multirow{2}{*}{\shortstack{MoCA\\SR $\uparrow$}} \\ 

Model  & Flow & RGB  &   &  & &  & & \\
\midrule
\multicolumn{6}{l}{\textit{\textbf{Flow-only methods}}} \\
\midrule
$^{\dag}$SIMO~\cite{Lamdouar21} & $\checkmark$  & \xmark & $-$ & $-$  & $-$ &    $67.8$ &  $62.0$  &  $-$  & $0.566$ \\
$^{\dag}$Meunier et al.~\cite{meunier2023unsupervised} & $\checkmark$  & \xmark &  $-$  &  $-$  & $-$ &   $73.2$ &  $55.0$  &  $59.4$  & $-$ \\
$^{\dag}$OCLR~\cite{Xie22} & $\checkmark$  & \xmark   & $46.5$  & $54.5$ &  $-$ & $72.1$ & $67.6$ & $70.0$ & $0.599$ \\
OCLR-real~\cite{Xie22} & $\checkmark$  & \xmark & $49.5$ & $55.7$ &  $-$  & $73.3$ & $65.9$ & $70.5$ & $0.605$ \\
\textbf{\flowsam~(seq, ViT-B) } & $\checkmark$  & \xmark &  $51.9$ &  $60.0$ & $-$ & $\mathbf{78.4}$  & $66.9$ & $69.0$ & $\mathbf{0.615}$  \\
\textbf{\flowsam~(seq, ViT-H)}& $\checkmark$  & \xmark  & $\mathbf{53.8}$ & $\mathbf{61.5}$ & $-$ & $78.0$ & $\mathbf{67.7}$ & $\mathbf{71.5}$  &  $0.604$ \\
\midrule
\multicolumn{6}{l}{\textit{\textbf{RGB-based methods}}} \\
\midrule
UnOVOST~\cite{luiten2020unovost} & \xmark & $\checkmark$  &  $-$ & $-$ & $66.4$ & $-$ & $-$ & $-$ & $-$  \\
Propose-Reduce~\cite{lin2021video} & \xmark & $\checkmark$  &  $-$ & $-$  &  $67.0$ & $-$ & $-$ & $-$ & $-$ \\
OCLR-flow~\cite{Xie22} + SAM~\cite{kirillov2023segment}  & $\checkmark$ & $\checkmark$ & $57.0$ & $62.0$   & $-$ & $80.6$ & $71.5$ & $79.2$ & $-$   \\
PMN~\cite{pmn}  & $\checkmark$ & $\checkmark$  & $-$ & $-$ & $-$ & ${85.6}$  &  $-$  &  ${77.8}$ & $-$  \\
Xie et al.~\cite{xie2023appearancebased} + SAM~\cite{kirillov2023segment} & $\checkmark$ & $\checkmark$   & $71.1$ & $70.9$ & $-$ & $86.6$ & $\mathbf{81.3}$  & $\mathbf{85.7}$  & $-$   \\
DEVA~\cite{cheng2023tracking} & \xmark & $\checkmark$  & $-$ & $-$  & $70.4$ & $87.6$ & $-$ & $-$ & $-$  \\
UVOSAM~\cite{zhang2024uvosam} & \xmark & $\checkmark$  & $-$ & $-$  & $\mathbf{77.5}$ & $ $ & $-$ & $-$ & $-$  \\
\textbf{\motionsam+\flowsam~(seq)} & $\checkmark$ & $\checkmark$   & $\mathbf{74.7}$ & $\mathbf{74.3}$ & $71.0$ & $\mathbf{87.7}$  & $80.1$ & $82.8$ &   $\mathbf{0.647}$  \\
\bottomrule
\end{tabular}}
\vspace{0.15cm}
\caption{\textbf{Sequence-level comparison on video object segmentation benchmarks.} ``${\dag}$'' indicates models that are trained without human annotations. ``seq'' indicates that our sequence-level predictions with object masks matched across frames. We adopt ~\motionsam~and~\flowsam~(ViT-H) to obtain the results in the last row.
}
\label{tab:vos_seq}
\end{table}

\vspace{3pt}
\noindent \textbf{Sequence-Level Performance.} 
For flow-based segmentation, we apply the mask association technique introduced in~\cref{sec:temp} to obtain sequence-level predictions from per-frame~\flowsam~results. 
To ensure a fair comparison, we additionally finetune the synthetic-trained OCLR~\cite{Xie22} model on the real-world dataset (DAVIS) with groundtruth annotations provided, resulting in ``OCLR-real'' results. As shown by the top part of Table \ref{tab:vos_seq},~\flowsam~(seq) demonstrates superior performance against OCLR-real, benefiting from the robust prior knowledge in pre-trained SAM.

For RGB-based segmentation, we obtain our sequence-level predictions by ~\flowsam+\motionsam. 
As shown in the lower part of Table~\ref{tab:vos_seq}, our method achieves outstanding performance across single- and multi-object benchmarks.
Note, DAVIS17 annotations are class-based, which could be unclear and inconsistent with the class-agnostic unsupervised VOS methods. More discussion and results are provided in~\Cref{supsec:quantitative}.

%% file: sec/5b-results.tex
\begin{figure}[t!]
    \centering
    \includegraphics[width=0.95\linewidth]{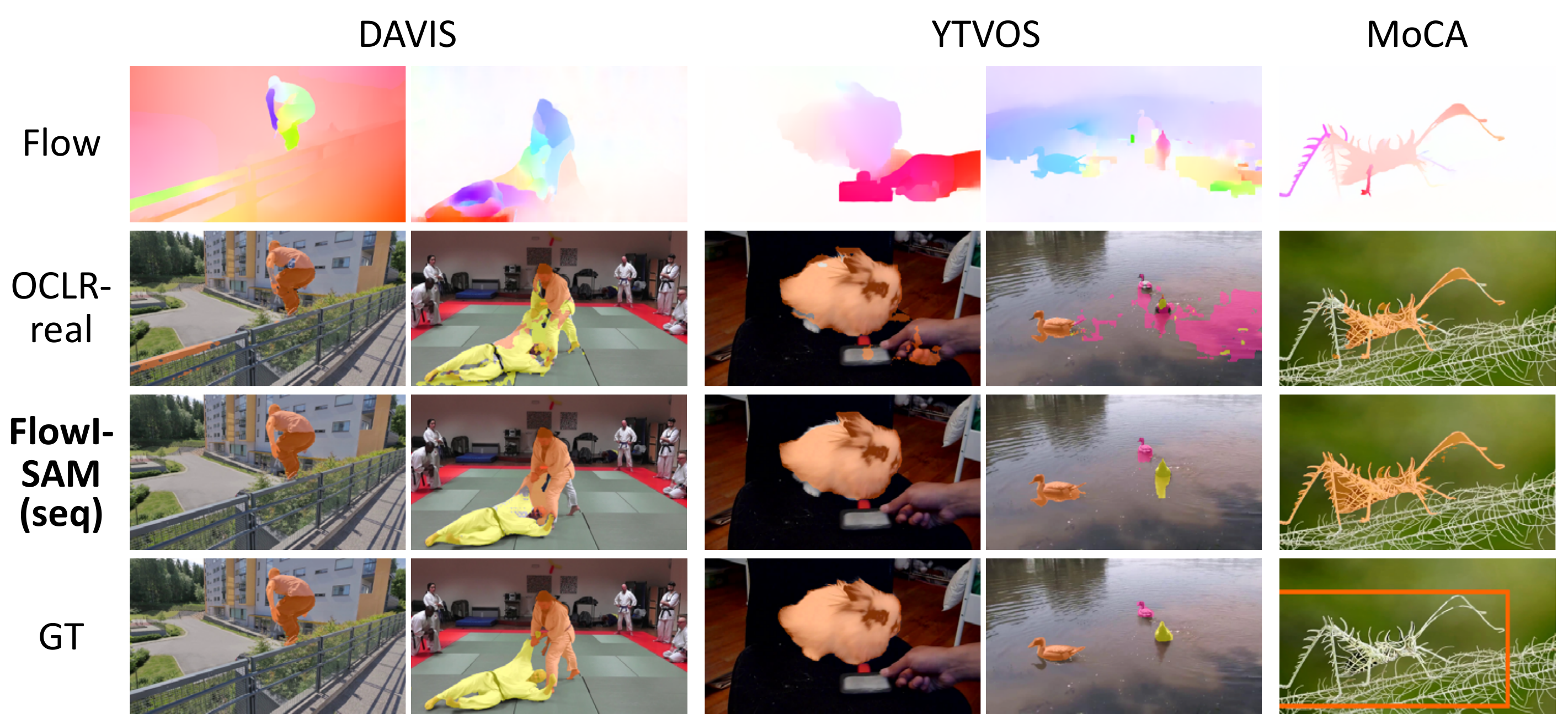}
    \vspace{-0.15cm}
    \caption{\textbf{Qualitative comparison of flow-only segmentation methods} on DAVIS (left), YTVOS (middle), and MoCA (right) sequences. Our~\flowsam~(seq) successfully identifies moving objects from noisy optical flow background ({\em e.g.,} the ducks in the fourth column). 
    }
    \label{fig:flowonlyvis}
\end{figure}

\begin{figure}[t!]
    \centering
    \includegraphics[width=0.95\linewidth]{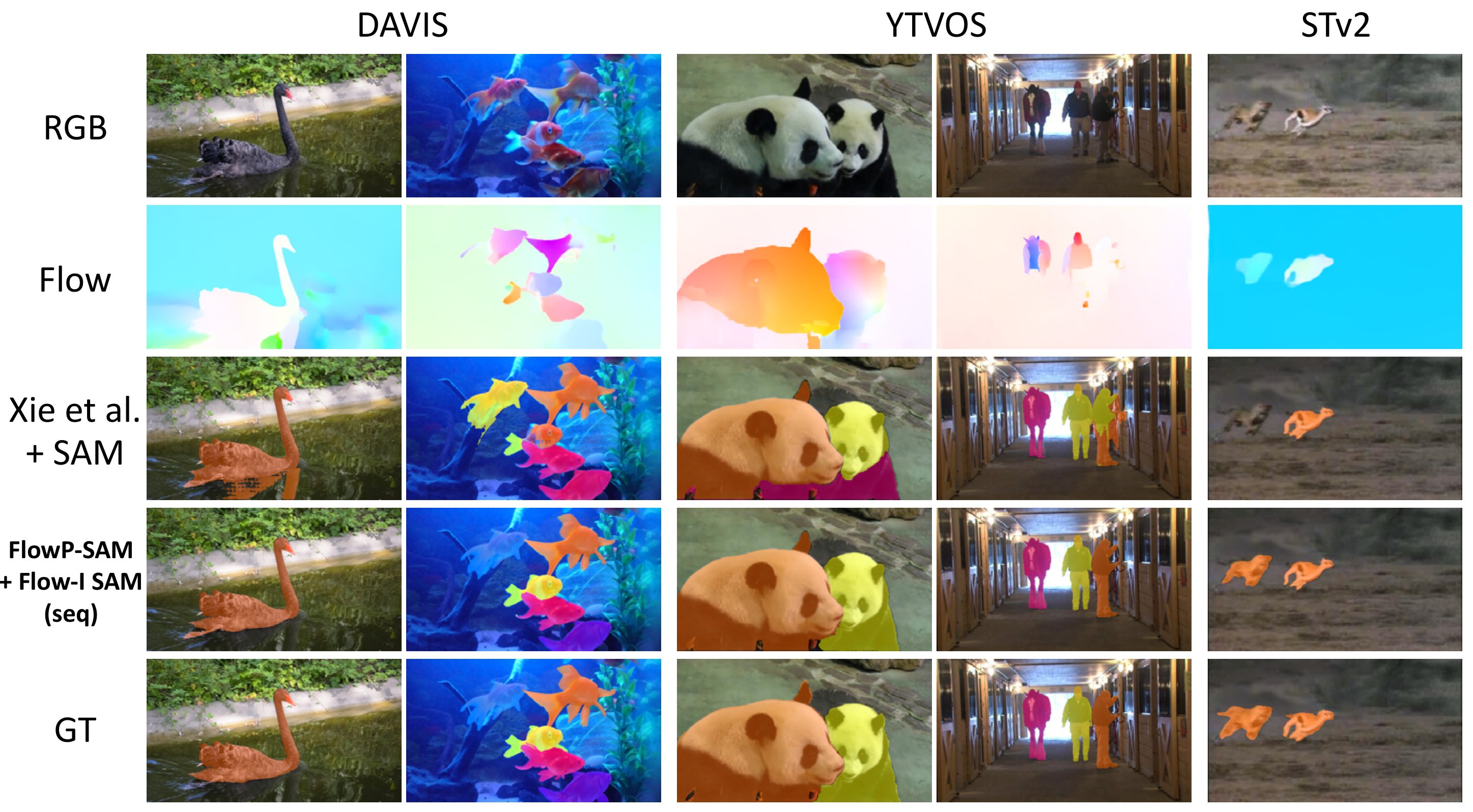}
    \vspace{-0.15cm}
    \caption{\textbf{Qualitative comparison of RGB-based segmentation methods} on DAVIS (left), YTVOS (middle), and SegTrackv2 (right). While the previous method (the third row) struggles to disentangle multiple moving objects ({\em e.g.,} mixed gold fishes in the second column), our~\motionsam+\flowsam~(seq) accurately separates and segments all moving objects. 
    }
    \label{fig:rgbbasedvis}
\end{figure}

\subsection{Qualitative Visualisations}
In this section, example visualisations are provided across multiple datasets. We refer more visualisations to~\cref{supsec:qualitative}.

\cref{fig:flowonlyvis} illustrates the segmentation predictions based on only optical flow inputs. Compared to OCLR-real, our~\flowsam~accurately identifies and disentangles the moving objects from the noisy backgrounds ({\em e.g.,}~ the person in the first column and the ducks in the fourth column), as well as extracts fine structures ({\em e.g.,}~the camouflaged insect in the fifth column) from optical flow.

\cref{fig:rgbbasedvis} further provides the visualisations of the RGB-based method, where the prior work (Xie et al.~\cite{xie2023appearancebased} + SAM~\cite{kirillov2023segment}) sometimes fails to (i) identify the moving objects ({\em e.g.,} missing leopard in the fifth column); (ii) distinguish between multiple objects ({\em e.g.,} entangled object segmentation in the second and fourth columns),
while our~\flowsam+\motionsam~(seq) incorporates RGB-based prediction with flow prompts, resulting in the accurate localisation and segmentation of moving objects.

%% file: sec/6-conclusion.tex
\section{Discussion}

In this paper, we focus on moving object segmentation in real-world videos, by incorporating per-frame SAM with motion information (optical flow) in two ways: (i) for flow-only segmentation, we introduce~\flowsam~that directly takes in optical flow as inputs; (ii) for RGB-based segmentation (\motionsam), we utilise motion information to generate flow prompts as guidance. The former (\flowsam) is particularly effective in scenarios with predominant motion and/or where RGB information might introduce confusion, such as in moving object detection and camouflaged object discovery. Additionally, owing to simple texture and a low cross-domain gap in optical flow, it generalises to diverse domains beyond everyday videos. On the other hand,~\motionsam~focuses on common videos where both RGB and motion are informative and can be utilised to resolve ambiguities for independent objects in common motion.

Both approaches deliver state-of-the-art performance in frame-level segmentation across single- and multi-object benchmarks. 
Additionally, we develop a frame-wise association method that amalgamates predictions from~\flowsam~ and ~\motionsam, achieving sequence-level segmentation predictions that outperform existing methods on DAVIS16, DAVIS17-m, YTVOS18-m, and MoCA.

The major limitation of this work is its extended running time, attributed to the computationally heavy image encoder in the vanilla SAM. However, our approach is generally applicable to other prompt-based segmentation models. With the emergence of more efficient versions of SAM, we anticipate a significant reduction in inference time.
\label{sec:conclusion}

%% file: sec/Acknowledgement.tex
\subsubsection{Acknowledgments.}
This research is supported by the UK EPSRC CDT in AIMS (EP/S024050/1), a Clarendon Scholarship, a Royal Society Research Professorship RP$\backslash$R1$\backslash$191132, and the UK EPSRC Programme Grant Visual AI (EP/T028572/1).

%% file: supp_insert.tex
\appendix
\begin{center}
    {\large \textbf{Appendix}}
\end{center}
\vspace{10pt} 

\noindent This Appendix consists of the following sections: \\\\
(i) In~\cref{supsec:impl}, we provide \textbf{implementation details} regarding architectural designs and experimental settings;  \\\\
(ii) In~\cref{supsec:ablation}, we conduct comprehensive \textbf{ablation studies} for both frame-level and sequence-level segmentation; \\\\
(iii) In~\cref{suppsec:combination} (\textbf{combining motion segmentation with motion-guided segmentation}), we discuss the rationale of combining results from \flowsam~and \motionsam~into a single prediction with some example cases; \\\\
(iv) In~\cref{supsec:quantitative}, additional \textbf{quantitative comparisons} are provided, along with a discussion on evaluating unsupervised object discovery using DAVIS17. \\\\
(v) In~\cref{supsec:qualitative}, \textbf{visualisations and failure cases} are discussed. \\\\
% \revised{(iv) In~\cref{supsec:difference_between_models}, we provide a discussion regarding the \textbf{differences between~\flowsam~and~\motionsam;}}
% (v) In~\cref{supsec:quantitative}, additional \textbf{quantitative comparisons} are provided;
% (vi) In~\cref{supsec:qualitative}, \textbf{visualisations and failure cases} are discussed.

\input{supsec/s1-implementation}

\clearpage
\input{supsec/s2-ablation}

\clearpage
\input{supsec/comb_ip_sam}

% \input{supsec/Difference_between_two_models}
\input{supsec/s3-quantitative}

% \clearpage
\input{supsec/s4-visualisation}

%% file: supsec/s1-implementation.tex
% \vspace{3pt} 
\section{Implementation Details}
\label{supsec:impl}
In this section, we summarise the experimental settings in our models, including hyperparameter choices, architecture details, and training datasets, with the detailed settings for frame-level and sequence-level segmentation separately discussed. The official code will be released upon acceptance.
% For more information regarding detailed architectures, hyperparameter settings, and sequence-level mask associations, please refer to the Supplementary Material.

\subsection{Frame-Level Segmentation}
\vspace{3pt} 
\label{supsec:impl_framelevel}

% \vspace{3pt} 
\par{\noindent \textbf{Training and Evaluation Datasets.}} 
For both~\flowsam~and~\motionsam, there are two major training stages: a synthetic pre-training on the simulated dataset proposed by~\cite{Xie22}, followed by finetuning on the real DAVIS sequences. These two datasets are adopted owing to their predominantly moving objects in their training sequences. A more detailed summary of the training datasets and corresponding evaluation benchmarks is listed in~\Cref{suptab:datasets}. For evaluation, apart from the  DAVIS validation sequences, we assess the zero-shot performance on YTVOS18-m, STv2, FBMS, and MoCA datasets. Notably, owing to occasional multi-object sequences in STv2 and FBMS, we evaluate the model that is trained on multi-object DAVIS sequences.

\begin{table}[thb!]
  \centering
  \vspace{-0.3cm}
  % \resizebox{\textwidth}{!}{
  \begin{tabular}{ccc}  
    \toprule
    \setlength\tabcolsep{10pt}
    \;Model \;& \;\;\;\;\;\;Training datasets \;\;\;\;\;\;& Evaluation datasets   \\
    \midrule
     \multirow{2}[2]{*}{\flowsam} & Syn+DAVIS16 &  DAVIS16, \textit{MoCA}\\ 
     \cmidrule(lr){2-3}
     & Syn+DAVIS17-m & DAVIS17-m, \textit{YTVOS18-m}, \textit{STv2}, \textit{FBMS} \\ 
     % & & \\
    \midrule
    \midrule
    % & & \\
     \multirow{3}[4]{*}{\motionsam}  &  Syn+DAVIS16 &  DAVIS16, \textit{MoCA}\\ 
     \cmidrule(lr){2-3}
     & Syn+DAVIS17 & DAVIS17, \textit{YTVOS18-m}, \textit{STv2}, \textit{FBMS} \\ 
     \cmidrule(lr){2-3}
     & Syn+DAVIS17-m & DAVIS17-m \\  
    \bottomrule
  \end{tabular}
  % }
  \vspace{0.15cm}
  \caption{\textbf{
  Training datasets and corresponding evaluation benchmarks.} Datasets in \textit{italic} indicate zero-shot generalisation during evaluation.}
  \label{suptab:datasets}
  \vspace{-0.3cm}
\end{table}

\vspace{10pt} 
\par{\noindent \textbf{Hyperparameter Settings.}} 
Regarding the input resolutions, we follow the default SAM settings to pad and resize the images to $1024 \times 1024$. After the frozen SAM encoder, the resultant dense spatial features are of size $64 \times 64$, with feature dimensions at $256$. During training of~\flowsam, we adopt a loss factor $\lambda_f=0.01$ for fIoU loss. For~\motionsam, we set both loss factors ($\lambda_f$ for fIoU loss and $\lambda_f$ for MOS loss) to $0.01$.

\vspace{6pt} 
\par{\noindent \textbf{Architectural Details.}} 
For~\flowsam, we preserve the architecture of the original SAM and directly finetune it with optical flow inputs. For~\motionsam, as detailed in the main text, a new flow prompt generator is introduced, with the details provided in~\cref{supalg:arch}.
% \az{It does not specify the transformer parameters}
\input{figs/architecture_code}

\vspace{6pt} 
\par{\noindent \textbf{Inference time analysis.}} 
As mentioned in~\cref{sec:conclusion} in the main text, the heavy image encoder in SAM leads to extended running times. We conduct a detailed inference time analysis and list the results of our default~\flowsam~and \motionsam~in the first row of~\Cref{suptab:runtime}. We also report the potential speed-ups of our method when adapting more efficient image encoders from more recent SAM-like models.
\begin{table}[h]
\vspace{-0.3cm}
\centering
\setlength\tabcolsep{4pt}
\begin{tabular}{cc|cc}  
\toprule
 \multicolumn{2}{c}{\flowsam} & \multicolumn{2}{c}{\motionsam}  \\

\cmidrule(r){1-2}
\cmidrule(r){3-4}
  \multirow{2}{*}{Encoder (Flow)} & \multirow{2}{*}{\shortstack{Running time \\ \emph{s/img}}} &  \multirow{2}{*}{Encoder (Flow $\mid$ RGB)} & \multirow{2}{*}{\shortstack{Running time \\ \emph{s/img}}}\\
& & & \\
\midrule
SAM-B& $0.46$ & SAM-B $\mid$  SAM-H & $0.97$  \\
\midrule
EfficientSAM-S~\cite{xiong2023efficientsam} & $0.07$ & EfficientSAM-S~\cite{xiong2023efficientsam} (both)   &  $0.08$    \\
SAM2-B+~\cite{ravi2024sam2} & $0.08$ &  SAM2-B+~\cite{ravi2024sam2} $\mid$ SAM2-L~\cite{ravi2024sam2}&  $0.16$  \\
\bottomrule
\end{tabular}
\vspace{0.15cm}
\caption{\textbf{Inference time analysis.} Our models can benefit from more efficient image encoders, with the potential speed-ups shown in last two rows.}
\label{suptab:runtime}
\vspace{-0.3cm}
\end{table}

\subsection{Sequence-Level Association}

In this section, we present the code snip used for sequence-level association (\cref{supalg:asso}).
{
\scriptsize
\input{figs/pseudocode}
}

%% file: figs/architecture_code.tex
% \vspace{-0.2cm}
\begin{algorithm}[htb!]
% \vspace{0.2cm}
\caption{Pseudo Code for Flow Prompt Generator}\label{supalg:arch}
\vspace{-0.2cm}
\begin{flushleft}
\scriptsize
\begin{alltt}
\textcolor{OliveGreen}{# Number of frame gaps g = 4 (for 1,-1,2,-2)}
\textcolor{OliveGreen}{# Image resolution H = W = 1024; Image channel size C = 3}
\textcolor{OliveGreen}{# Feature resolution h = w = 64; Feature channel size c = 256}

\textcolor{Orange}{""" Flow feature encoding """}
f = SAM_image_encoder(F)      \textcolor{OliveGreen}{# b g C H W (input flow F)}
                              \textcolor{OliveGreen}{# b g c h w (dense flow features f)}
f = Average(f)                \textcolor{OliveGreen}{# b c h w, averaging over frame gaps}

\textcolor{Orange}{""" Point Prompt encoding """}
pp = SAM_prompt_encoder(p)    \textcolor{OliveGreen}{# b 2 (point coordinates p)}
                              \textcolor{OliveGreen}{# b c (point prompt token pp)}
                              
\textcolor{Orange}{""" Query initialisation """}
mos = Embedding(b, c)         \textcolor{OliveGreen}{# b c (learnable moving object score token mos)}
fp = Embedding(b, 4, c)       \textcolor{OliveGreen}{# b 4 c (4 learnable flow prompt tokens fp)}
q = Concat(mos, fp, pp)       \textcolor{OliveGreen}{# b 6 c, concatenating all tokens to form queries q}

\textcolor{Orange}{""" Flow transformer """}
for i in range(2):            \textcolor{OliveGreen}{# Two layers}
    \textcolor{OliveGreen}{# A standard transformer decoder layer with query, key, value inputs}
    \textcolor{OliveGreen}{# Feed-forward layer dimension = 512; Number of heads = 8}
    q = transformer_decoder_layer(q, f, f)   \textcolor{OliveGreen}{# b 6 c}
    
mos, fp, _ = q                \textcolor{OliveGreen}{# Output tokens}

\textcolor{OliveGreen}{# Output flow prompt token fp (b 4 c) will be injected into the segmentation module.}

\textcolor{OliveGreen}{# A three-layer MLP with a sigmoid function as the last activation function}
\textcolor{OliveGreen}{# Feed-forward layer dimension = 256}
mos_score = MLP(mos)          \textcolor{OliveGreen}{# b 1 (moving object score estimation mos_score)}
\end{alltt}
\end{flushleft}
\vspace{-0.2cm}
\end{algorithm}
% \vspace{-0.2cm}
% \vspace{-0.5cm}

%% file: figs/pseudocode.tex
% \begin{verbatim}
% def threeway_hungarian(m1, m2, m3):
%     ious_23 = iou(m2, m3)
%     _, idx_23 = linear_sum_assignment(-ious_23)
%     m3_aligned = m3[idx_23]
%     ious_13 = iou(m1, m3_aligned)
%     ious_12 = iou(m1, m2)
%     _, idx_13 = linear_sum_assignment(-ious_13)
%     _, idx_12 = linear_sum_assignment(-ious_12)
%     return m2[idx_12], (idx_12 == idx_13)

% def temp_consistency(p, m, b1, b2, f1, f2):
%     m_aligned, c1 = threeway_hungarian(p, m, b1)
%     _, c2 = threeway_hungarian(p, m, b2)
%     _, c3 = threeway_hungarian(p, m, f1)
%     _, c4 = threeway_hungarian(p, m, f2)
%     c = ((c1+c2+c3+c4)/4 >= 0.5)
%     return m_aligned * c + p * (1-c)    
% \end{verbatim}

% \begin{alltt}
% \scriptsize
% def threeway_hungarian(m1, m2, m3):
%     ious_23 = iou(m2, m3)
%     _, idx_23 = linear_sum_assignment(-ious_23)
%     m3_aligned = m3[idx_23]
%     ious_13 = iou(m1, m3_aligned)
%     ious_12 = iou(m1, m2)
%     _, idx_13 = linear_sum_assignment(-ious_13)
%     _, idx_12 = linear_sum_assignment(-ious_12)
%     return m2[idx_12], (idx_12 == idx_13)

% def temp_consistency(p, m, b1, b2, f1, f2):
%     m_aligned, c1 = threeway_hungarian(p, m, b1)
%     _, c2 = threeway_hungarian(p, m, b2)
%     _, c3 = threeway_hungarian(p, m, f1)
%     _, c4 = threeway_hungarian(p, m, f2)
%     c = ((c1+c2+c3+c4)/4 >= 0.5)
%     return m_aligned * c + p * (1-c)    
% \end{alltt}

% \vspace{-0.2cm}
\begin{algorithm}[htb!]
% \vspace{0.2cm}
\caption{Pseudo Code for Sequence-Level Association}\label{supalg:asso}
\vspace{-0.2cm}
\begin{flushleft}
\small%scriptsize
\begin{alltt}
def threeway_hungarian(m1, m2, m3):
    ious_23 = iou(m2, m3)
    _, idx_23 = linear_sum_assignment(-ious_23)
    m3_aligned = m3[idx_23]
    ious_13 = iou(m1, m3_aligned)
    ious_12 = iou(m1, m2)
    _, idx_13 = linear_sum_assignment(-ious_13)
    _, idx_12 = linear_sum_assignment(-ious_12)
    return m2[idx_12], (idx_12 == idx_13)

def temp_consistency(p, m, b1, b2, f1, f2):
    m_aligned, c1 = threeway_hungarian(p, m, b1)
    _, c2 = threeway_hungarian(p, m, b2)
    _, c3 = threeway_hungarian(p, m, f1)
    _, c4 = threeway_hungarian(p, m, f2)
    c = ((c1+c2+c3+c4)/4 >= 0.5)
    return m_aligned * c + p * (1-c)    
\end{alltt}
\end{flushleft}
\vspace{-0.2cm}
\end{algorithm}
% \vspace{-0.2cm}

%% file: supsec/s2-ablation.tex
\section{Ablation Study}
\label{supsec:ablation}

\subsection{Frame-Level Segmentation:~\flowsam}

\vspace{2pt}
\par{\noindent \textbf{Optical Flow Frame Gaps.}}
As demonstrated in~\Cref{tab:abla_flowgap}, utilizing optical flow inputs with multiple frame gaps {({\em i.e.,}~$1$,-$1$,$2$,-$2$)} results in noticeable performance boosts across both multi- and single-object benchmarks. This improvement is attributed to the consistency of motion information over extended temporal ranges, which effectively mitigates the impact of noise in optical flow inputs caused by slow movements, partial motions, {\em etc.}

\begin{table}[htpb!]
  \centering
  \vspace{-0.3cm}
  \begin{tabular}{ccc}  
    \toprule
    \setlength\tabcolsep{10pt}
    \multirow{2}{*}{\;\;\shortstack{Optical flow \\ frame gaps}\;\;} & \multirow{2}{*}{\;\;\shortstack{DAVIS17-m \\ $\mathcal{J} \uparrow$}\;\;}  & \multirow{2}{*}{\;\;\shortstack{DAVIS16 \\ $\mathcal{J} \uparrow$}\;\;}   \\
     & & \\
    \midrule
    $1,$ -$1$ & $64.5$ & $78.0$ \\ 
    $2,$ -$2$ & $65.3$ & $77.7$ \\ 
    $1,$-$1,2,$-$2$  & $65.7$ &  $79.1$\\ 
    \bottomrule
  \end{tabular}
  \vspace{0.15cm}
  \caption{\textbf{
  The frame gaps of input flows to~\flowsam.} The SAM ViT-H image encoder is adopted. The results are shown for frame-level predictions.}
  \label{tab:abla_flowgap}
  \vspace{-0.2cm}
\end{table}

\vspace{2pt}
\par{ \noindent \textbf{Combination of Flow Features.}}
We have explored two combination schemes: (i) taking the maximum; and (ii) averaging across different frame gaps. According to~\Cref{tab:abla_flowcomb}, the averaging approach yields superior results.

\begin{table}[htpb!]
  \centering
  \vspace{-0.3cm}
  \begin{tabular}{ccc}  
    \toprule
    \setlength\tabcolsep{10pt}
    \multirow{2}{*}{\;\shortstack{Flow features  \\ combination}\;} & \multirow{2}{*}{\;\;\shortstack{DAVIS17-m \\ $\mathcal{J} \uparrow$}\;\;}  & \multirow{2}{*}{\;\;\shortstack{DAVIS16 \\ $\mathcal{J} \uparrow$}\;\;}   \\
     & & \\
    \midrule
    Taking max & $65.0$ & $78.2$ \\ 
    Averaging  & $65.7$ &  $79.1$ \\ 
    \bottomrule
  \end{tabular}
  \vspace{0.15cm}
   \caption{\textbf{The combination of dense flow features in~\flowsam.} The SAM ViT-H image encoder is adopted. The results are shown for frame-level predictions.}
   \label{tab:abla_flowcomb}
   \vspace{-0.2cm}
\end{table}

\vspace{2pt}
\par{\noindent \textbf{Optical Flow Estimation Method.}} To ensure a fair comparison, we follow prior works~\cite{Yang21a,Xie22} and apply RAFT~\cite{Teed20} as the flow estimation method. Table~\ref{suptab:abla_flow} demonstrates how the input flow quality affects the segmentation results, which also verifies our default choice of RAFT for flow estimation.

\begin{table}[htpb!]
  \centering
  \vspace{-0.3cm}
  \begin{tabular}{ccc}  
    \toprule
    \setlength\tabcolsep{10pt}
    \multirow{2}{*}{\;\;\shortstack{Optical flow \\ method}\;\;} & \multirow{2}{*}{\;\;\shortstack{DAVIS17-m \\ $\mathcal{J} \uparrow$}\;\;}  & \multirow{2}{*}{\;\;\shortstack{DAVIS16 \\ $\mathcal{J} \uparrow$}\;\;}   \\
     & & \\
    \midrule
    MaskFlownet~\cite{zhao2020maskflownet} & $61.7$ & $76.1$ \\ 
    RAFT~\cite{Teed20}  & $65.7$ &  $79.1$\\ 
    \bottomrule
  \end{tabular}
  \vspace{0.15cm}
  \caption{ 
  % \az{Shouldn't the title be Comparison of optical flow computation methods for~\flowsam.} ?  \junyu{sorry. I modified it.}
  \textbf{Comparison of optical flow methods.} The results are predicted by frame-level~\flowsam, where the SAM ViT-H image encoder is adopted to encode optical flow with frame gaps $\{1$,-$1$,$2$,-$2\}$. We adopt RAFT as the default optical flow estimation method.}
  \label{suptab:abla_flow}
  \vspace{-1.4cm}
\end{table}

\begin{figure}[t]
    \centering
    \includegraphics[width=\linewidth]{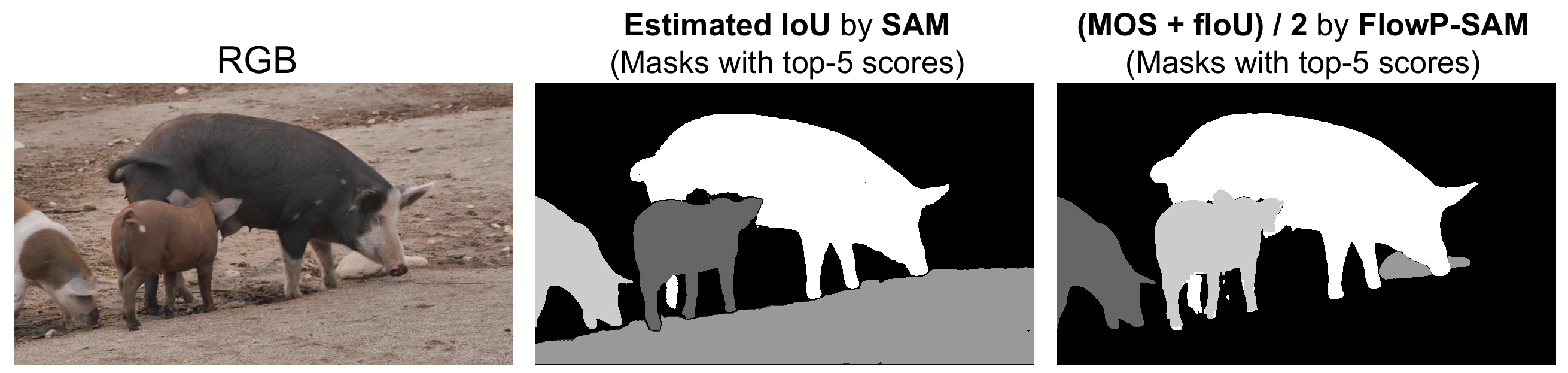}
    \vspace{-0.6cm}
    \caption{\textbf{Qualitative comparison between IoUs estimated by SAM and scores predicted by~\motionsam\ for selecting masks of foreground objects.} Only masks with top-5 scores are shown. The masks with higher scores are brighter and arranged to the front.} 
    \vspace{-0.3cm}
    \label{supfig:score_comp}
\end{figure}

\subsection{Frame-Level Segmentation:~\motionsam}
\vspace{3pt} 
\par{\noindent \textbf{Comparison with SAM~\cite{kirillov2023segment}.}}
Since the vanilla SAM is not trained to explicitly identify moving foreground objects, for a fair comparison, we consider an alternative setup by specifying the objects with their groundtruth centroids as point prompt inputs. 
We apply this setting to both SAM and~\motionsam, with the performance summarised in~\Cref{suptab:abla_sam}.

Notably, the original SAM formulation employs four mask tokens to generate masks at different semantic levels, including default, sub-parts, parts, and whole. For object-level segmentation, we examine the performance of the $1^{\text{st}}$ (default) and $4^{\text{th}}$ (whole) output mask channels. According to~\Cref{suptab:abla_sam}, while the $4^{\text{th}}$ output channel shows a superior performance in SAM, we observe an opposite trend for~\motionsam, where the $1^{\text{st}}$ mask token yields better results. Therefore, the $1^{\text{st}}$ output mask token is adopted as our default setting.

\begin{table}[thbp!]
  \centering
  \vspace{-0.4cm}
  \begin{tabular}{ccccc}  
    \toprule
    \setlength\tabcolsep{6pt}
    \multirow{2}{*}{Model} & \multirow{2}{*}{\;\;\shortstack{Output mask \\ token/channel}\;\;}& \multirow{2}{*}{\;\shortstack{DAVIS17-m \\ $\mathcal{J} \uparrow$}\;}  &  \multirow{2}{*}{\;\;\shortstack{DAVIS17 \\ $\mathcal{J} \uparrow$}\;\;} & \multirow{2}{*}{\;\;\shortstack{DAVIS16 \\ $\mathcal{J} \uparrow$}\;\;}   \\
     & & & & \\
    \midrule
    SAM~\cite{kirillov2023segment} & $1^{\text{st}}$& $46.6$ &  $48.3$ & $42.7$\\
    SAM~\cite{kirillov2023segment}  & $4^{\text{th}}$ & $71.4$ &  $68.0$ & $73.6$ \\ 
    ~\motionsam~ & $4^{\text{th}}$ & $79.5$ &  $72.6$ & $82.9$ \\ 
    \midrule
    ~\motionsam~ & $1^{\text{st}}$ & $80.0$ &  $73.4$ & $85.6$ \\ 
    \bottomrule
  \end{tabular}
  \vspace{0.15cm}
  \caption{\textbf{Quantitative comparison between SAM and~\motionsam.} The centroid of each groundtruth object mask is provided as a point prompt. The results are shown for frame-level predictions, and our default~\motionsam~utilises the $1^{\text{st}}$ mask token.}
  \label{suptab:abla_sam}
  \vspace{-0.7cm}
\end{table}

Furthermore, our proposed scores (MOS and fIoU) serve as more accurate indicators of complete foreground objects, compared to the original IoU scores estimated by SAM. \cref{supfig:score_comp} illustrates a qualitative example, where simply ranking the masks with higher IoU scores would result in the selection of background masks (\emph{e.g.,} the ground) and incomplete masks (\emph{e.g.,} the middle pig with an ear missing). In contrast, the averaged MOS + fIoU scores help to correctly identify the \emph{complete foreground} object masks, which verifies our claim that the proposed scores capture the ``objectness'' of masks.

\vspace{6pt} 
\par{\noindent \textbf{Comparison with SAM2~\cite{ravi2024sam2}.}}
Following the comparison setup in~\cref{supfig:score_comp} (\emph{i.e.}, prompting by a single ground truth centroid), we compare our~\motionsam~with more recent SAM2. As shown in~\Cref{suptab:sam2}, though SAM2 gets noticeably improved from SAM, our~\motionsam~still demonstrates superior object segmentation performance.

\begin{table}[h!]
\vspace{-0.4cm}
  \centering
  \setlength\tabcolsep{6pt}
  \begin{tabular}{cccc}  
    \toprule
    \multirow{2}{*}{Model} & \multirow{2}{*}{\;\shortstack{DAVIS17-m \\ $\mathcal{J} \uparrow$}\;}  &  \multirow{2}{*}{\;\;\shortstack{DAVIS17 \\ $\mathcal{J} \uparrow$}\;\;} & \multirow{2}{*}{\;\;\shortstack{DAVIS16 \\ $\mathcal{J} \uparrow$}\;\;}   \\
     & & & \\
    \midrule
    SAM~\cite{kirillov2023segment}  &  $71.4$ &  $68.0$ &  $73.6$\\
    SAM2~\cite{ravi2024sam2} &  $79.4$  & $69.7$ & $84.0$  \\ 
    \midrule
     ~\motionsam~ & $80.0$  & $73.4$ & $85.6$ \\ 
    \bottomrule
  \end{tabular}
  \vspace{0.15cm}
  \caption{\textbf{Quantitative comparison across SAM, SAM2, and~\motionsam.} The centroid of each groundtruth object mask is provided as a point prompt. The best results among the four output channels are shown.}
  \label{suptab:sam2}
   \vspace{-0.6cm}
\end{table}

% \vspace{3pt} 
\par{\noindent \textbf{Comparison with flow-based method + SAM~\cite{kirillov2023segment}.}} An alternative method for moving object segmentation is to apply SAM to refine the flow-predicted masks (\emph{e.g.}, by OCLR-flow and by~\flowsam). However, such a two-stage method prevents the interaction between RGB and flow information, leading to inferior performance compared to the end-to-end~\motionsam, as can be seen in \Cref{suptab:abla_altermethod}.

\begin{table}[htpb!]
\vspace{-0.2cm}
  \centering
  \begin{tabular}{ccc}  
    \toprule
    \setlength\tabcolsep{10pt}
    \multirow{2}{*}{\;\;Method\;\;} & \multirow{2}{*}{\;\;\shortstack{DAVIS17-m \\ $\mathcal{J} \uparrow$}\;\;}  & \multirow{2}{*}{\;\;\shortstack{DAVIS16 \\ $\mathcal{J} \uparrow$}\;\;}   \\
     & & \\
    \midrule
    OCLR-flow~\cite{Xie22} + SAM~\cite{kirillov2023segment}  &  $62.0$ &   $80.6$ \\
    \flowsam~+ SAM~\cite{kirillov2023segment} & $72.6$ & $82.4$ \\ 
    \midrule
    \motionsam & $78.5$ &  $86.1$\\ 
    \bottomrule
  \end{tabular}
  \vspace{0.15cm}
  \caption{\textbf{Comparison with two-stage methods.} The results are shown for frame-level predictions.}
  \label{suptab:abla_altermethod}
  \vspace{-0.3cm}
\end{table}

\vspace{3pt} 
\par{\noindent \textbf{Number of Transformer Decoder Layers.}} We further investigate the effects of the number of transformer decoder layers in the flow prompt generator (in~\motionsam). As can be observed from~\Cref{suptab:abla_ptranslayer}, increasing the layer number from $2$ (default) to $4$ does not contribute to a noticeable performance change.
 
\begin{table}[hptb!]
\vspace{-0.3cm}
  \centering
  \begin{tabular}{cccc}  
    \toprule
    \setlength\tabcolsep{6pt}
    \multirow{2}{*}{\;\shortstack{Transformer \\ decoder layers}\;} & \multirow{2}{*}{\;\shortstack{DAVIS17-m \\ $\mathcal{J} \uparrow$}\;}  &  \multirow{2}{*}{\;\;\shortstack{DAVIS17 \\ $\mathcal{J} \uparrow$}\;\;} & \multirow{2}{*}{\;\;\shortstack{DAVIS16 \\ $\mathcal{J} \uparrow$}\;\;}   \\
     & & \\
    \midrule
    $2$ & $78.5$ &  $69.9$ & $86.1$\\
    $4$ & $79.2$ &  $69.1$ & $85.1$ \\ 
    \bottomrule
  \end{tabular}
  \vspace{0.15cm}
  \caption{\textbf{
  The number of layers in the transformer decoder of flow prompt generator in~\motionsam.} The results are shown for frame-level predictions, and by default, we adopt $2$ transformer decoder layers.}
  \label{suptab:abla_ptranslayer}
  \vspace{-0.9cm}
\end{table}

\vspace{2pt}
\par{\noindent \textbf{Optical Flow Estimation Method.}} 
Different from~\flowsam~results, 
% (shown in~\Cref{suptab:abla_flow}), 
\Cref{suptab:abla_flow_for_flowpsam} demonstrates a lower impact of flow quality on~\motionsam. This is because in~\motionsam, the optical flow is mainly adopted to prompt the region of moving objects, while its boundary accuracy only affects the final results marginally.

\begin{table}[h!]
\vspace{-0.4cm}
  \centering
  \begin{tabular}{ccc}  
    \toprule
     \setlength\tabcolsep{10pt}
    \multirow{2}{*}{\;\;\shortstack{Optical flow \\ method}\;\;} & \multirow{2}{*}{\;\;\shortstack{DAVIS17-m \\ $\mathcal{J} \uparrow$}\;\;}  & \multirow{2}{*}{\;\;\shortstack{DAVIS16 \\ $\mathcal{J} \uparrow$}\;\;}   \\
     & & \\
    \midrule
    MaskFlownet~\cite{zhao2020maskflownet} & $65.3$ & $78.9$ \\ 
    RAFT~\cite{Teed20}  & $65.7$ &  $79.1$\\ 
    \bottomrule
  \end{tabular}
  \vspace{0.15cm}
  \caption{\textbf{Comparison of optical flow methods.} The results are predicted by frame-level~\motionsam, where the SAM ViT-B image encoder is adopted to encode optical flow with frame gaps $\{1$,-$1$,$2$,-$2\}$. We adopt RAFT as the default optical flow estimation method.}
  \label{suptab:abla_flow_for_flowpsam}
  \vspace{-1.2cm}
\end{table}

\subsection{Sequence-Level Mask Association}

In this section, we conduct ablation studies on our sequence-level mask association module. We show comparison against two baselines that constitute our method: (i) propagating the previous mask only, and (ii) Hungarian-matching between past and present masks only. %\az{of what to what?}. 
We also perform ablations to show the performance gains by averaging the confidence scores across different neighbouring frames as compared to picking one neighbouring frame. The results are shown in Table~\ref{tab:abs-seq}.

We observe that (i) Hungarian matching alone presents a strong baseline for RGB-based tracking (using \motionsam~+~\flowsam), where frame-level predictions are largely consistent; (ii) However, in flow-only cases (using \flowsam) where object identities might disappear, Hungarian matching is prone to lose track of the object. %,  but less so in Flow-only cases where object identities may disappear. This {\color{red}failure case of object disappearing} \az{what does this refer to? Objects that disappear?} 
This issue of losing track can be solved using our temporal consistency method, where choosing to propagate appropriate masks helps maintain object permanence. We observe that flow-only method (\flowsam) benefits considerably from this. We postulate further that using more frames may help even further, but this will involve computing new optical flows, whereas we only currently use those already adopted as input to \flowsam~and/or \motionsam.

\begin{table}[htpb!]
\centering
\vspace{-0.4cm}
\begin{tabular}{lcc}
\toprule & RGB-based & Flow-only \\
 & \;\; DAVIS17 $\mathcal{J}$ $\uparrow$ \;\;& \;\;DAVIS17-m $\mathcal{J}$ $\uparrow$ \;\; \\ \midrule
Propagation only & 34.9  & 29.3 \\
Hungarian only & 68.1 & 48.5 \\ 
Temporal consistency ($-1$) & 69.0 & 48.7 \\
Temporal consistency ($\pm1$) & 69.4 & 51.7 \\
\midrule
Temporal consistency ($\pm1$, $\pm2$) & 71.0 & 61.5 \\ \bottomrule
\end{tabular}
\vspace{0.15cm}
\caption{\textbf{Ablation study on sequence-level mask association.} The frame-wise masks are predicted by \motionsam~+~\flowsam~(ViT-H) for RGB-based, and \flowsam~(ViT-H) for flow-only segmentation, with different methods for sequence-level mask association.}% \az{Add the methods that are used here.}}
\label{tab:abs-seq}
\vspace{-0.7cm}
\end{table}

%% file: supsec/comb_ip_sam.tex
\section{Combining Motion Segmentation with Motion-Guided Segmentation}
\label{suppsec:combination}
%\charig{maybe it doesn't need a separate section, but putting it here first as it doens't seem to belong to the current list}
In the main paper, we have quantitatively shown that combining the results of \flowsam~and \motionsam~simply by layering the latter behind the former yields better results. Here, we discuss the reason why this is the case, and also provide visualisation examples.

We observe that one of the main failure modes of \motionsam~(and other RGB-based methods) is when the model completely fails to identify the object due to poor appearance, such as occlusion, camouflage, motion blur, small object size, or bad lighting conditions. In these cases, \flowsam~(and other flow-only methods) shine as they are agnostic to appearance and objectness.

In such cases, layering the motion segmentation masks behind RGB-based segmentation masks is a very simple and sensible solution. Specifically, we concatenate the \flowsam~predictions behind that of \motionsam, followed by removing any overlaps. In this way, the regions predicted by both models will always belong to \motionsam.

\begin{figure}[hbt!]
    \centering
    \vspace{-0.3cm}
    \includegraphics[width=1\linewidth]{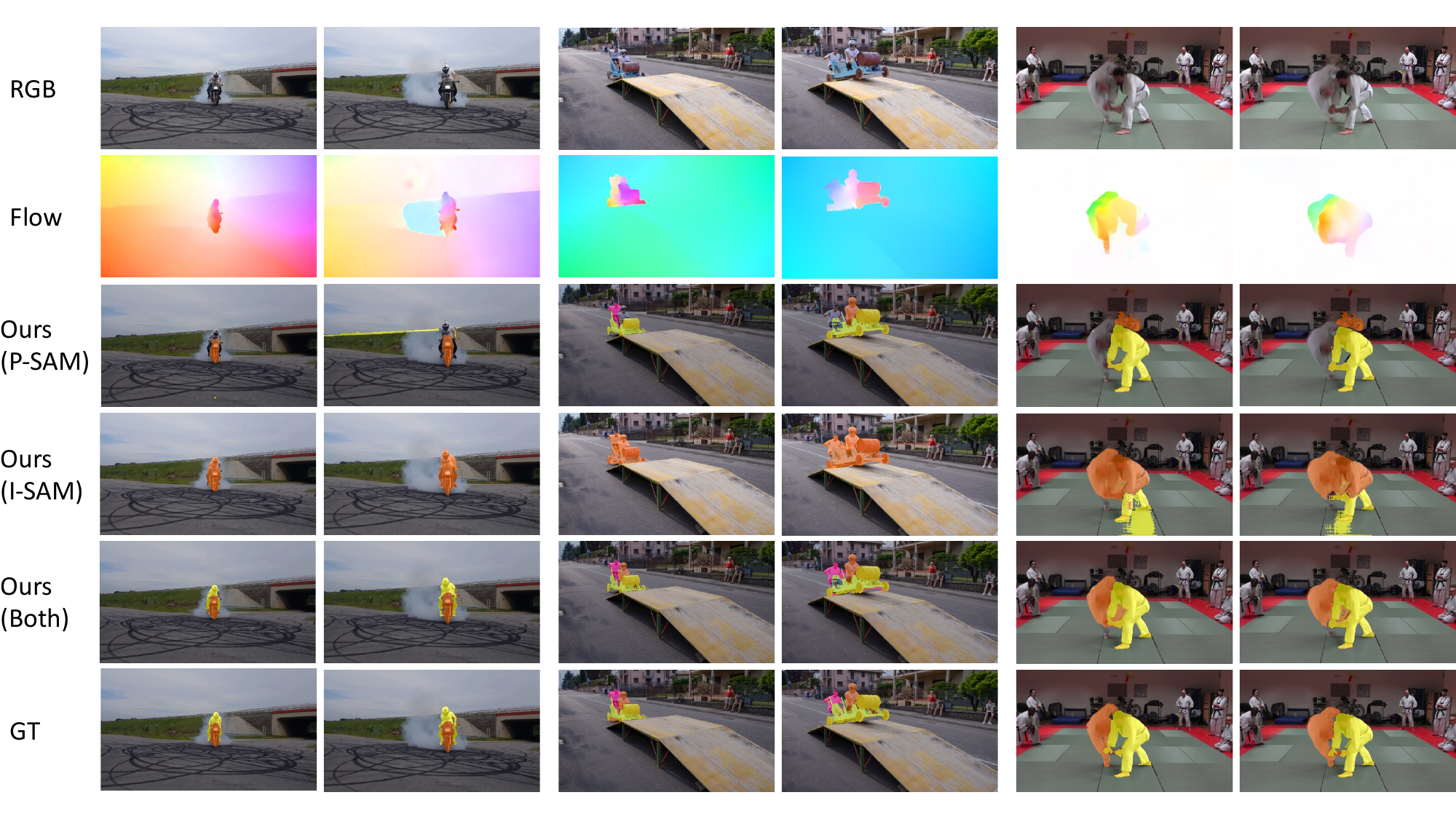}
    \vspace{-0.7cm}
    \caption{\textbf{Combining \flowsam~and \motionsam.} This example shows combining the two predictions by layering \flowsam's prediction behind \motionsam~allows recovery of lost objects undetected by \motionsam~due to small object (left), occlusion (middle), and motion blur (right). Note that both models individually make wrong predictions.
    }
    \label{supfig:combination}
    \vspace{-0.2cm}
\end{figure}

We show the effectiveness of this method in Figure \ref{supfig:combination}. In each example case, the predicted mask from \motionsam~missed an object, whereas the predicted mask from \flowsam~grouped all moving objects as a single object. We show that layering these two masks together allows the prediction from \flowsam~to `fill in' the gap that \motionsam~failed to predict. Notably, the object identities are also correctly separated. This is because \flowsam~, being layered behind, does not over-segment regions that are already segmented by \motionsam.

%Additionally, in cases of two objects moving together and the RGB-based method is only able to detect one of them, the layering also helps with separation of the two objects, even in cases where both \flowsam~and \motionsam~individually make wrong predictions. %\az{clarify the previous sentence -- individual predictions from what? }

%Evidently in the quantitative results in the main paper, this results in higher accuracy overall, and we show examples that lead to these improvements in Figure \ref{supfig:combination}. {\color{red} This also suggests another use-case for motion segmentation as a task -- which is to directly complement RGB-based video segmentation models when they fail.} \az{not clear what this last sentence means: why does it motivate another case? what is existing about `anomaly detection' or surveillance?}
%In general, this also allows motion segmentation to be a useful complementary model to image segmentation, which motivates another use case of this task on top of the existing ones (camouflage, anomaly detection, surveillance, etc.)  \charig{I wanted to say that motion segmentation has many useful applications already (camouflage etc) -- this paper suggests one more which is to complement RGB-based methods in post-processing.}

%We observe that the combined results are better mostly in the case where \motionsam~completely fails to detect an object. Heavy reliance on appearance resulted in the model not segmenting some objects including those under motion blur and small objects. On the other hand, motion segmentation is object-agnostic, and allows for the correct segmentation in these cases. This justifies the reason for layering the motion segmentation behind image segmentation as a safety net in case the latter fails. We show some qualitative results in Figure \charig{figure}.

%We also considered switching the layering, and found that this did not yield good results. This is mainly because \flowsam~failure cases of either over-segmenting (objects moving causing background to also move) or under-segmenting (objects partially moving) do not get recovered by simply layering another segmentation mask behind.

%% file: supsec/s3-quantitative.tex
\section{Quantitative Results}
\label{supsec:quantitative}
For multi-object segmentation, we report the performance on both IoU ($\mathcal{J}$) and contour accuracy ($\mathcal{F}$), with \Cref{subtab:multivos_frame} and \Cref{subtab:multivos_seq} comparing across frame-level and sequence-level methods, respectively.

\begin{table}[tbph!]
\centering
\vspace{-0.4cm}
\setlength\tabcolsep{2.5pt}
\resizebox{\textwidth}{!}{
\begin{tabular}{lccccccccccc}  
\toprule
  &  & & \multicolumn{3}{c}{YTVOS18-m} & \multicolumn{3}{c}{DAVIS17-m} & \multicolumn{3}{c}{DAVIS17} \\ 
\cmidrule(lr){4-6}
\cmidrule(lr){7-9}
\cmidrule(lr){10-12}
{} & \; Flow &  \; RGB \; &
$\mathcal{J} \& \mathcal{F}$ $\uparrow$ &  $\mathcal{J}$ $\uparrow$ &  \; $\mathcal{F}$ $\uparrow$   \; &
$\mathcal{J} \& \mathcal{F}$ $\uparrow$ &  $\mathcal{J}$ $\uparrow$ &  \;$\mathcal{F}$ $\uparrow$  \;  &
$\mathcal{J} \& \mathcal{F}$ $\uparrow$ &  $\mathcal{J}$ $\uparrow$ &  \;$\mathcal{F}$ $\uparrow$  \;  \\
%Model  & Flow & RGB  &   &  & &  & & \\
\midrule
\multicolumn{6}{l}{\textit{\textbf{Flow-only methods}}} \\
\midrule
$^{\dag}$MG~\cite{Yang21a} & $\checkmark$ & \xmark &$33.3$ & $37.0$ & $29.6$ &  $35.8$ & $38.4$  &  $33.2$ &  $-$  &  $-$  & $-$  \\ 
\textbf{\flowsam~(ViT-B)} & $\checkmark$ & \xmark  & $53.9$& $56.7$& $51.2$& $62.3$& $63.2$& $61.3$&  $-$  &  $-$  & $-$  \\ 
\textbf{\flowsam~(ViT-H)} & $\checkmark$ & \xmark  &$\mathbf{56.4}$& $\mathbf{58.6}$& $\mathbf{54.2}$& $\mathbf{64.8}$& $\mathbf{65.7}$& $\mathbf{63.9}$&  $-$  &  $-$  & $-$  \\
% \textbf{\motionsam+\flowsam} & & &$0.0$& $0.0$& $0.0$& $0.0$& $0.0$& $0.0$& $0.0$& $0.0$& $0.0$  \\
\midrule
\multicolumn{6}{l}{\textit{\textbf{RGB-based methods}}} \\
\midrule
$^{\dag}$VideoCutLER~\cite{wang2023videocutler} & \xmark & $\checkmark$  & $57.0$& $59.0$& $55.1$& $57.3$& $57.4$& $57.2$& $43.6$& $41.7$& $45.5$  \\
$^{\dag}$Safadoust et al.~\cite{Safadoust23} & \xmark & $\checkmark$  &  $-$  &  $-$  & $-$ & $59.2$& $59.3$& $59.2$&  $-$  &  $-$  & $-$  \\
MATNet~\cite{zhou20} & $\checkmark$ & $\checkmark$  &  $-$  &  $-$  & $-$&  $-$  &  $-$  & $-$& $58.6$& $56.7$& $60.4$  \\
\textbf{\motionsam} &  $\checkmark$ & $\checkmark$  &$76.7$& $76.9$& $76.4$& $78.9$& $78.5$& $79.3$& $72.7$& $69.9$& $75.4$  \\  
\textbf{\motionsam+\flowsam} &  $\checkmark$ & $\checkmark$  &$\mathbf{77.4}$& $\mathbf{77.4}$& $\mathbf{77.5}$& $\mathbf{80.1}$& $\mathbf{80.0}$& $\mathbf{80.2}$& $\mathbf{74.6}$& $\mathbf{71.6}$& $\mathbf{77.6}$  \\
% \textbf{\flowsam~(seq, ViT-H)} & & & $0.0$& $0.0$& $0.0$& $0.0$& $0.0$& $0.0$& $0.0$& $0.0$& $0.0$  \\
% \textbf{\motionsam+\flowsam~(seq)} & & &$0.0$& $0.0$& $0.0$& $0.0$& $0.0$& $0.0$& $0.0$& $0.0$ & $0.0$ \\
\bottomrule
\end{tabular}}
\vspace{0.15cm}
\caption{\textbf{Frame-level comparison on multi-object segmentation benchmarks.} ``${\dag}$'' indicates models that are trained without human annotations. For the results in the last row, we combine frame-level predictions from~\motionsam~and~\flowsam~(ViT-H).
}
\vspace{-0.9cm}
\label{subtab:multivos_frame}
\end{table}

\begin{table}[!htb]
\centering
\vspace{-0.4cm}
\setlength\tabcolsep{2.5pt}
\resizebox{\textwidth}{!}{
\begin{tabular}{lccccccccccc}  
\toprule
  &  & & \multicolumn{3}{c}{YTVOS18-m} & \multicolumn{3}{c}{DAVIS17-m} & \multicolumn{3}{c}{DAVIS17} \\ 
\cmidrule(lr){4-6}
\cmidrule(lr){7-9}
\cmidrule(lr){10-12}
{} & \; Flow &  \; RGB \; &
$\mathcal{J} \& \mathcal{F}$ $\uparrow$ & $\mathcal{J}$ $\uparrow$ &  \;$\mathcal{F}$ $\uparrow$ \;&
$\mathcal{J} \& \mathcal{F}$ $\uparrow$ & $\mathcal{J}$ $\uparrow$ &\; $\mathcal{F}$ $\uparrow$\; &
$\mathcal{J} \& \mathcal{F}$ $\uparrow$ & $\mathcal{J}$ $\uparrow$ & \;$\mathcal{F}$ $\uparrow$  \;\\
%Model  & Flow & RGB  &   &  & &  & & \\
\midrule
\multicolumn{6}{l}{\textit{\textbf{Flow-only methods}}} \\
\midrule
$^{\dag}$OCLR~\cite{Xie22} & $\checkmark$  & \xmark  &   $45.3$ & $46.5$ & $44.1$ &  $55.1$ & $54.5$ & $55.7$ &  $-$  &  $-$  & $-$ \\
OCLR-real~\cite{Xie22} & $\checkmark$  & \xmark  & $47.5$ & $49.5$ & $45.5$ &  $56.2$ & $55.7$ & $56.7$  &  $-$  &  $-$  & $-$ \\
\textbf{\flowsam~(seq, ViT-B) } & $\checkmark$  & \xmark & $50.2$& $51.9$& $48.4$& $59.3$& $60.0$& $58.6$&  $-$  &  $-$  & $-$ \\
\textbf{\flowsam~(seq, ViT-H)}& $\checkmark$  & \xmark  &$\mathbf{52.1}$& $\mathbf{53.8}$& $\mathbf{50.3}$& $\mathbf{61.0}$& $\mathbf{61.5}$& $\mathbf{60.5}$& $-$  &  $-$  & $-$  \\
% \textbf{\motionsam+\flowsam} & & &$0.0$& $0.0$& $0.0$& $0.0$& $0.0$& $0.0$& $0.0$& $0.0$& $0.0$  \\
\midrule
\multicolumn{6}{l}{\textit{\textbf{RGB-based methods}}} \\
\midrule
UnOVOST~\cite{luiten2020unovost} & \xmark & $\checkmark$   & $-$  &  $-$  & $-$ &  $-$  &  $-$  & $-$ & $67.9$& $66.4$& $69.3$  \\
Propose-Reduce~\cite{lin2021video} & \xmark & $\checkmark$  &  $-$  &  $-$  & $-$&  $-$  &  $-$  & $-$ & $70.4$& $67.0$& $73.8$  \\
OCLR-flow~\cite{Xie22} + SAM~\cite{kirillov2023segment}  & $\checkmark$ & $\checkmark$  & $58.5$& $57.0$& $60.0$& $64.2$& $62.0$& $66.4$&  $-$  &  $-$  & $-$ \\
Xie et al.~\cite{xie2023appearancebased} + SAM~\cite{kirillov2023segment} & $\checkmark$ & $\checkmark$ &$70.6$& $71.1$& $70.2$& $71.5$& $70.9$& $72.1$&  $-$  &  $-$  & $-$  \\
DEVA~\cite{cheng2023tracking} & \xmark & $\checkmark$   & $-$  &  $-$  & $-$&  $-$  &  $-$  & $-$& $73.4$& $70.4$& $76.4$  \\
UVOSAM~\cite{zhang2024uvosam} & \xmark & $\checkmark$   & $-$  &  $-$  & $-$&  $-$  &  $-$  & $-$& $\mathbf{80.9}$& $\mathbf{77.5}$& $\mathbf{84.9}$  \\
\textbf{\motionsam+\flowsam~(seq)} & $\checkmark$ & $\checkmark$ &$\mathbf{75.2}$& $\mathbf{74.7}$& $\mathbf{75.7}$& $\mathbf{75.0}$& $\mathbf{74.3}$& $\mathbf{75.6}$& $73.6$& $71.0$ & $76.1$ \\
\bottomrule
\end{tabular}}
\vspace{0.15cm}
\caption{\textbf{Sequence-level comparison on multi-object segmentation benchmarks.} ``${\dag}$'' indicates models that are trained without human annotations. ``seq'' indicates that our sequence-level predictions with object masks matched across frames. We adopt~\motionsam~and~\flowsam~(ViT-H) to obtain the results in the last row. %{\color{blue}Those marked with * are slightly better than that reported in the Main paper, we will update.}
}
\vspace{-0.7cm}
\label{subtab:multivos_seq}
\end{table}

\vspace{2pt} 
\par{\noindent \textbf{Discussion on DAVIS17 annotations.}}
It is important to note that we found the DAVIS17 dataset may not be suitable for unsupervised VOS evaluations due to issues with object definitions. This can be understood in two aspects: (i) Ambiguous object definitions within the dataset due to class-based annotations. For example, a helmet on a cyclist is not considered a separate object, whereas a smartphone or handbag held by a person is. Such annotations pose a problem for our \emph{class-agnostic} object discovery method; (ii) Inconsistencies between annotations and motion information, where objects moving together are labeled separately based on their classes. This contradiction can confuse our model, as we use motion information as prompts for object discovery. This is also the underlying reason why we adopt DAVIS17-m (and YTVOS18-m) for evaluation, as they provide a clearer definition of objectness based on (joint) movements.

%% file: supsec/s4-visualisation.tex
\section{Qualitative Results and Failure Cases}
\label{supsec:qualitative}
\begin{figure}[hbt!]
    \centering
    \includegraphics[width=1\linewidth]{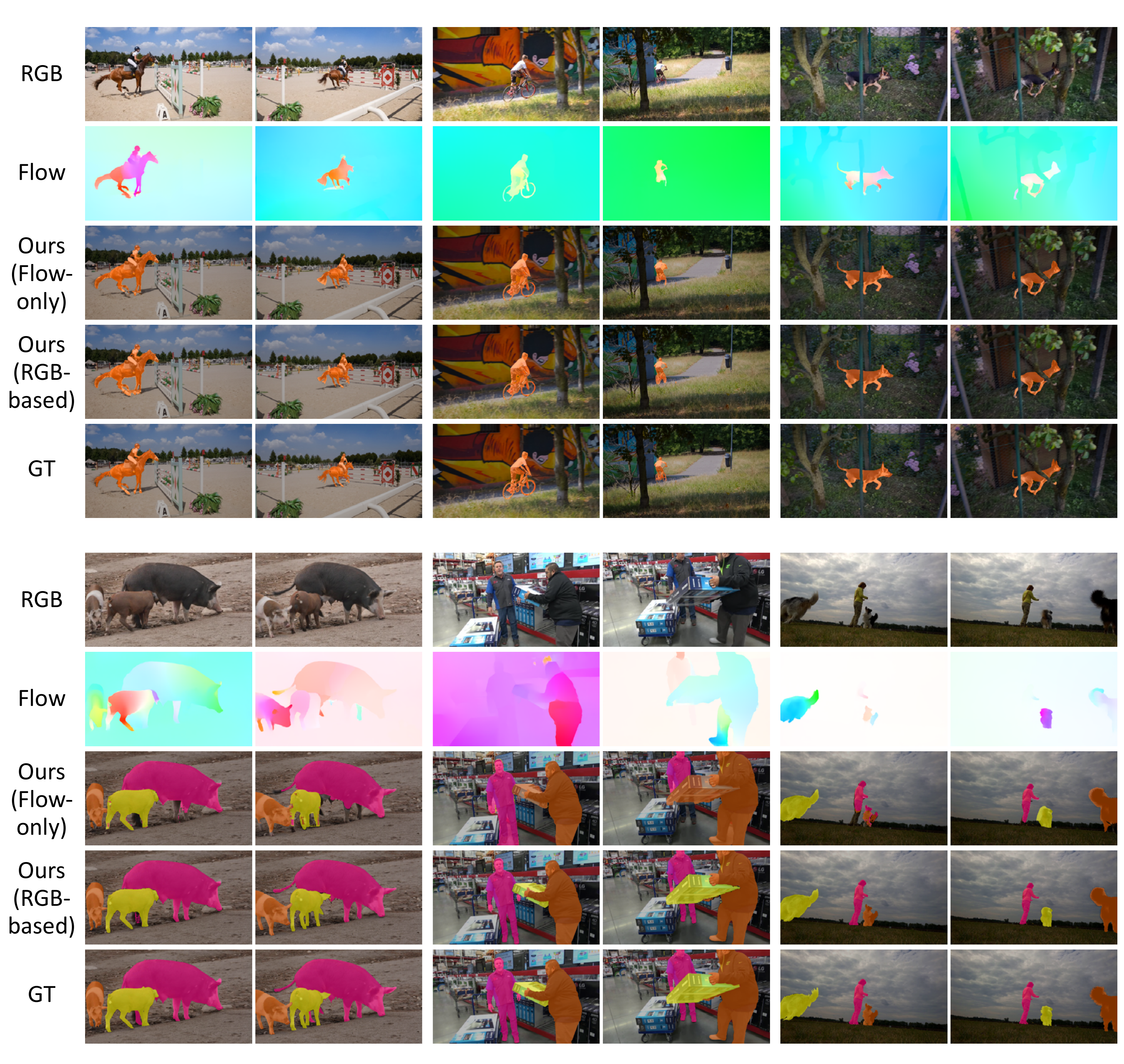}
    \vspace{-0.5cm}
    \caption{\textbf{Qualitative visualisation on DAVIS sequences.} The sequence-level predictions are shown, with~\flowsam~(seq) and~\motionsam~+~\flowsam~(seq) being our flow-only and RGB-based methods, respectively. Our flow-only method correctly identifies multiple moving objects based on noisy optical flow inputs ({\em e.g.,}~three pigs in the bottom left), while our RGB-based method yields more accurate segmentation masks.
    }
    \label{supfig:dvsvis}
    \vspace{-0.3cm}
\end{figure}

\begin{figure}[thbp!]
    \centering
    \includegraphics[width=1\linewidth]{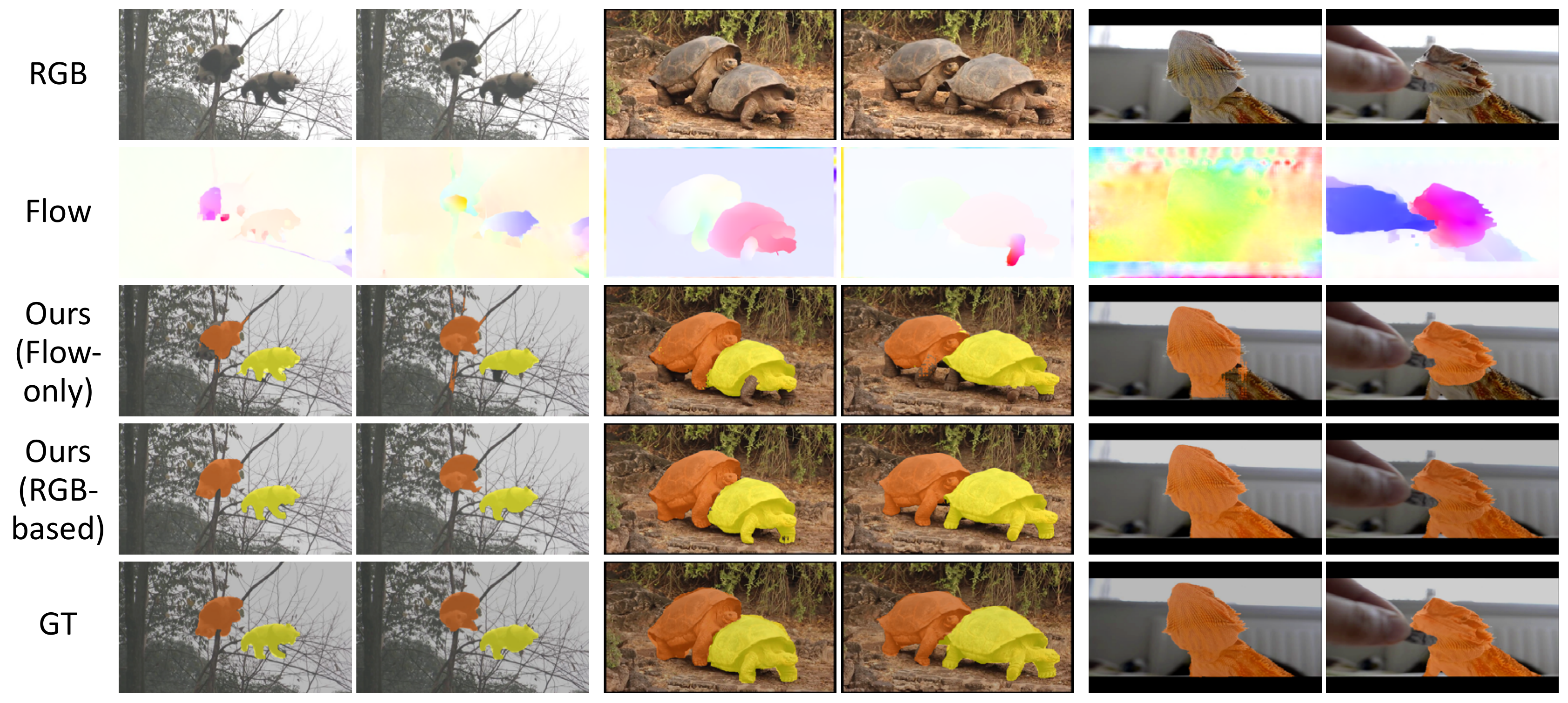}
    \vspace{-0.5cm}
    \caption{\textbf{Qualitative visualisation on YTVOS sequences.} The last sequence provides a partial motion example, where flow-only segmentation fails to recover the whole object mask.
    % The sequence-level predictions are shown, with~\flowsam~(seq) and~\motionsam~+~\flowsam~(seq) being our flow-only and RGB-based methods, respectively.
    }
    \label{supfig:ytvosvis}
    \vspace{-0.cm}
\end{figure}

\begin{figure}[hbpt!]
    \centering
    \includegraphics[width=1\linewidth]{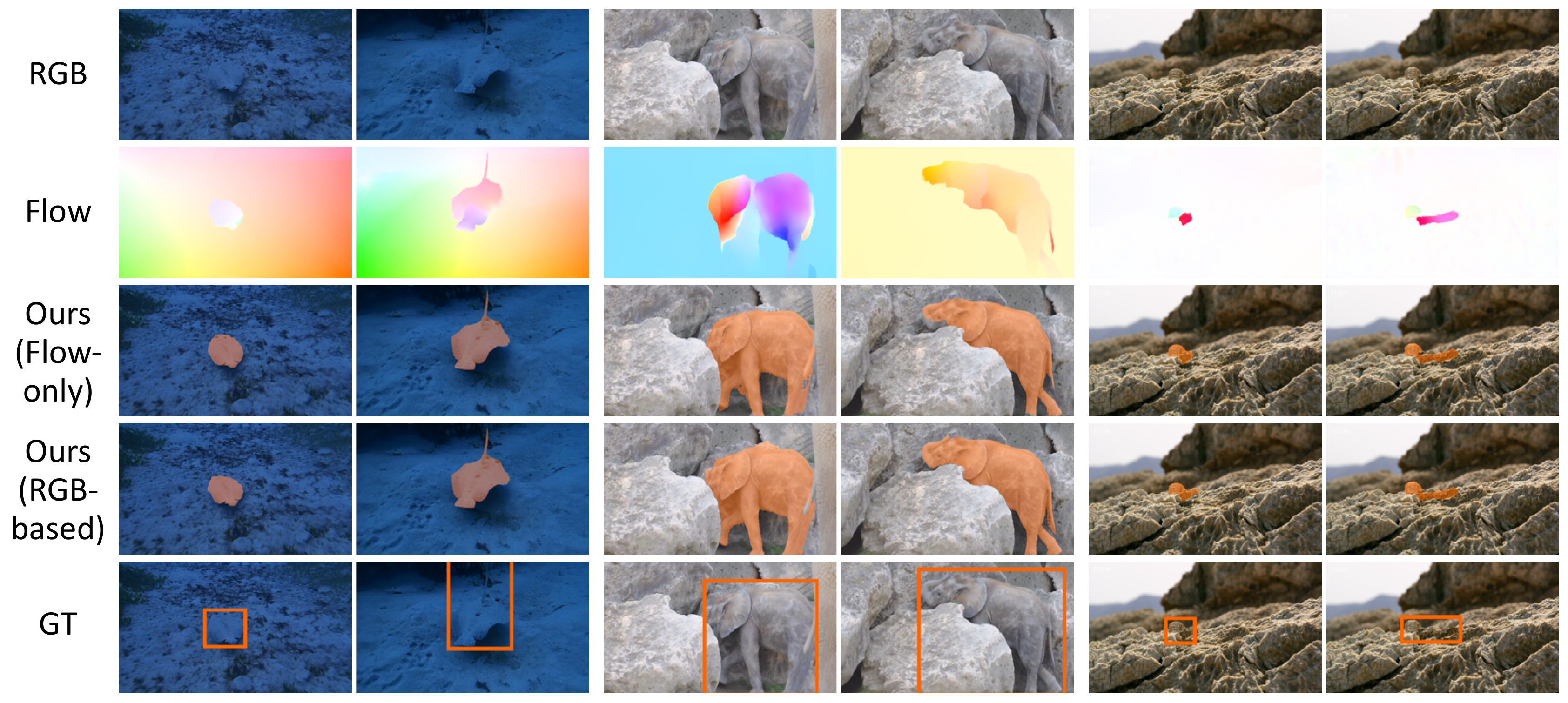}
    \vspace{-0.5cm}
    \caption{\textbf{Qualitative visualisation on MoCA sequences.} 
    % The sequence-level predictions are shown, with~\flowsam~(seq) and~\motionsam~+~\flowsam~(seq) being our flow-only and RGB-based methods, respectively.
    Both flow-only and RGB-based methods (with flow prompts) are capable of discovering the camouflaged object by leveraging motion information.
    }
    \label{supfig:mocavis}
    \vspace{-0.cm}
\end{figure}

\begin{figure}[hbtp!]
    \centering
    \includegraphics[width=1\linewidth]{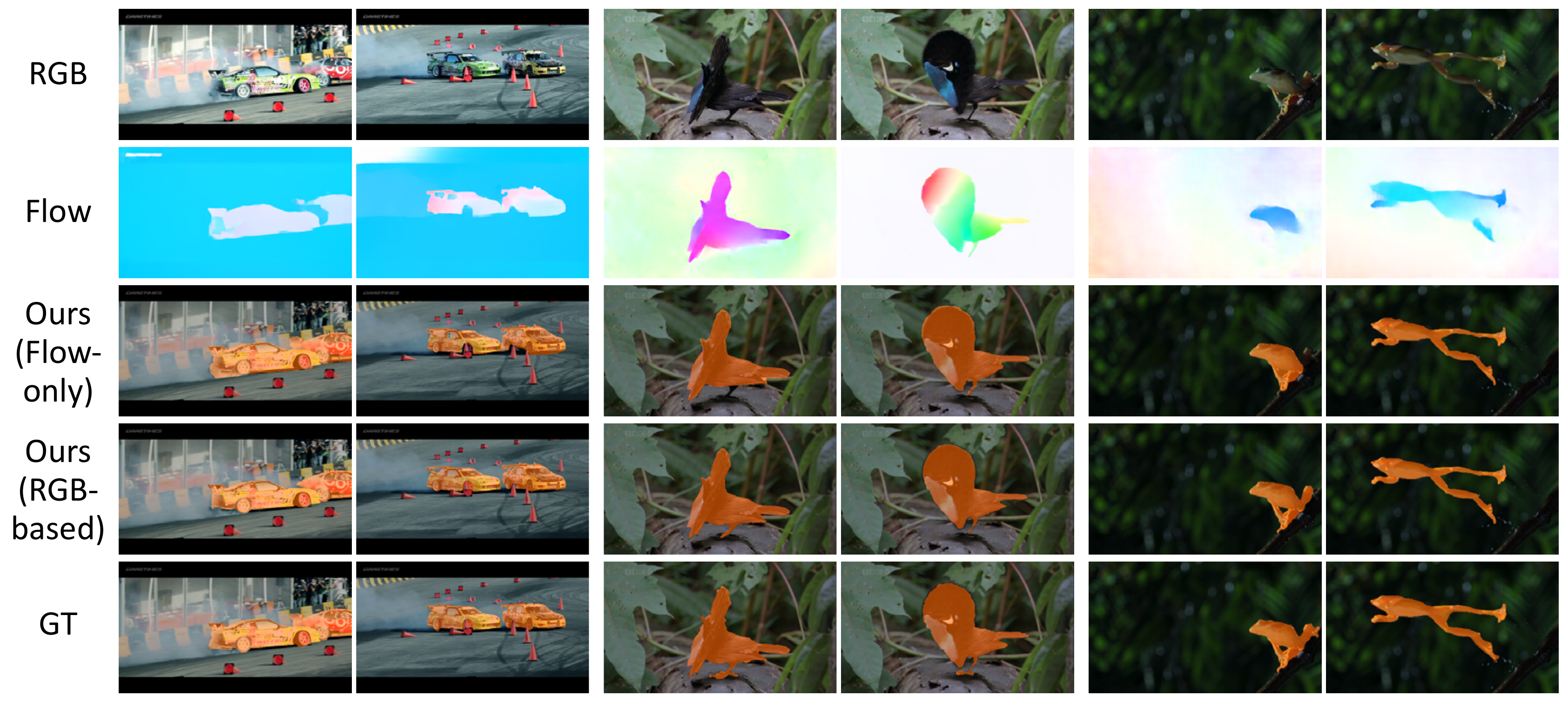}
    \vspace{-0.6cm}
    \caption{\textbf{Qualitative visualisation on STv2 sequences.} With the predominant object locomotion (therefore clean optical flow fields), both flow-only and RGB-based methods yield accurate segmentation masks.
    % The sequence-level predictions are shown, with~\flowsam~(seq) and~\motionsam~+~\flowsam~(seq) being our flow-only and RGB-based methods, respectively.
    }
    \label{supfig:stv2vis}
    \vspace{-0.4cm}
\end{figure}

\begin{figure}[hbtp!]
    \centering
    \includegraphics[width=1\linewidth]{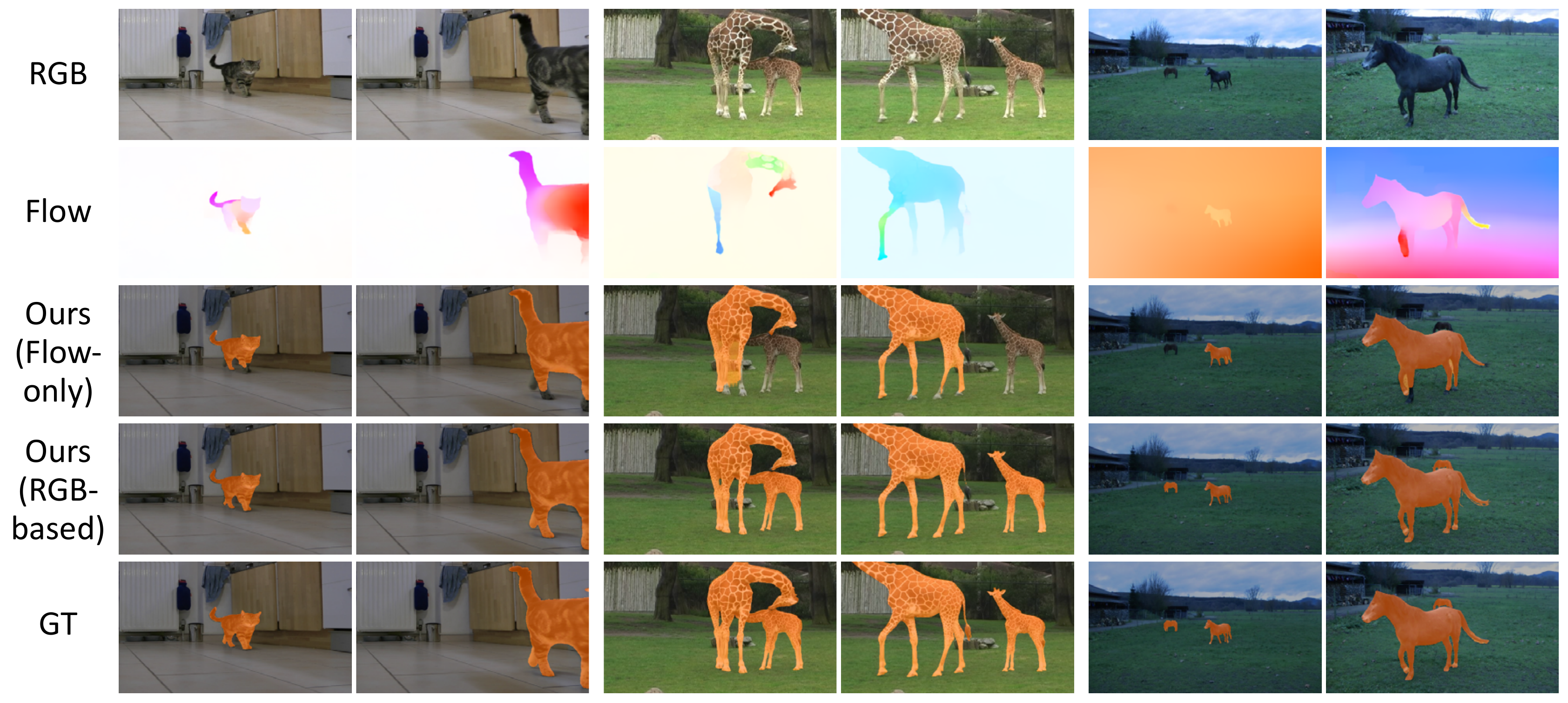}
    \vspace{-0.6cm}
    \caption{\textbf{Qualitative visualisation on FBMS sequences.} 
    % The sequence-level predictions are shown, with~\flowsam~(seq) and~\motionsam~+~\flowsam~(seq) being our flow-only and RGB-based methods, respectively.
    As shown in the second sequence ({\em i.e.,} the giraffes), the flow-only method is not capable of discovering the occasionally stationary object (the little giraffe), whereas the RGB-based method identifies both foreground giraffes correctly.
    }
    \label{supfig:fbmsvis}
    \vspace{-0.4cm}
\end{figure}

Additional visualisations are provided for our sequence-level segmentation results on various datasets, including DAVIS17 (\cref{supfig:dvsvis}), YTVOS18-m (\cref{supfig:ytvosvis}), MoCA (\cref{supfig:mocavis}), STv2 (\cref{supfig:stv2vis}), and FBMS (\cref{supfig:fbmsvis}). Note that, we adopt \flowsam~(seq) as the flow-only method, and~\motionsam~+~\flowsam~(seq) for RGB-based segmentation. 

\vspace{3pt} 
\par{\noindent \textbf{Failure Cases.}} For~\flowsam, one common failure case is related to the uninformative optical flow inputs. For instance, in the third sequence in~\cref{supfig:ytvosvis} and the second sequence in~\cref{supfig:fbmsvis}, (partially) stationary objects are not captured by the motion fields, therefore leading to missing objects/parts in the resultant flow-only segmentation.

\begin{figure}[pbth!]
    \centering
    \includegraphics[width=1\linewidth]{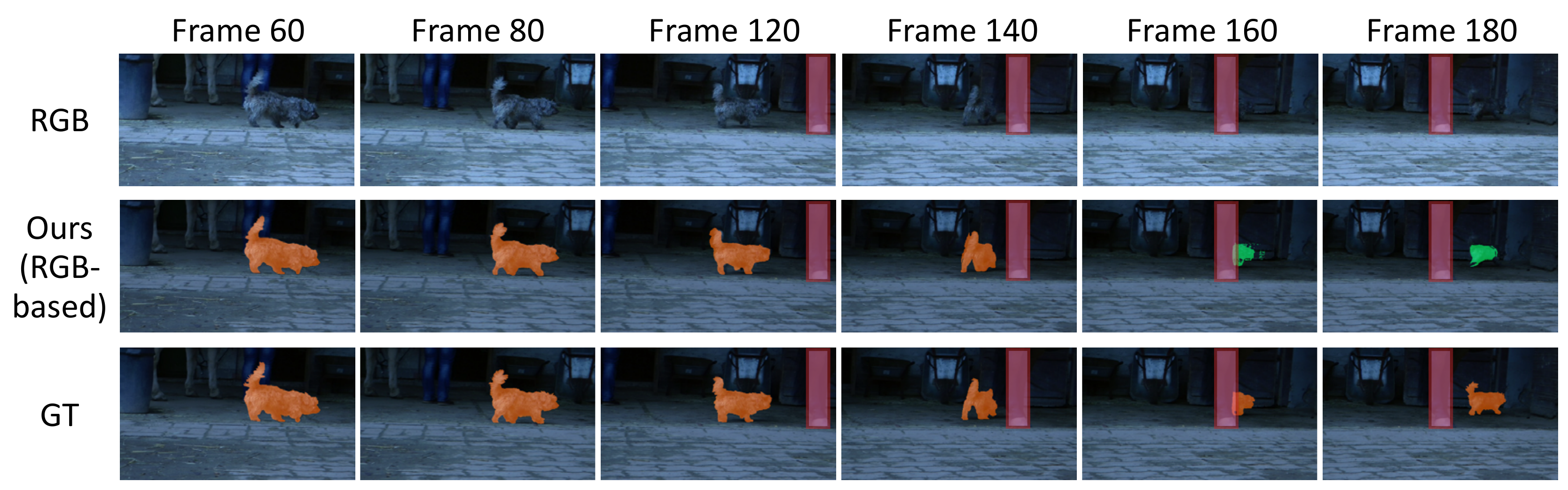}
    \vspace{-0.6cm}
    \caption{\textbf{A failure case.} This example sequence is from FBMS. For a clearer demonstration, the occluder (a horse foot) is labelled with the red box, and the object ID mis-matching is indicated by different colours (orange and green) of object masks.
    }
    \label{supfig:fbmsvisfail}
    \vspace{-0.4cm}
\end{figure}

Another limitation in this work is that the sequence-wise association fails in some cases, as shown in \cref{supfig:fbmsvisfail}. Here, the occlusion is long (10 frames), and the model has lost track of the object. This can indeed be improved by giving a longer temporal context in the temporal consistency, though our method does provide a strong baseline.
%In this example, the object identity has been lost as the object (in orange) moves behind an occluder (in red).
%During which, the frame-wise segmentation model is unable to find the occluded object (also partially due to poor lighting), hence the association module chooses to propagate the previous mask.
%However, the flow-propagated mask is `stuck' to the left of the occluder as it never gets warped across the occluder.
%After the object reappears, the frame-wise mask prediction (in green) no longer overlaps with the propagated mask, which did not cross the occluder. 
%This can be improved by preserving object identity throughout the sequence instead of autoregressively, though our method provides a strong and simple baseline. 
%\charig{maybe too much info?}

